\documentclass[nohyperref]{article}

\usepackage{microtype}
\usepackage{graphicx}
\usepackage{subfigure}
\usepackage{booktabs} 

\usepackage{hyperref}

 \usepackage[accepted]{icml2022_TAGML}

\usepackage{amsmath}
\usepackage{amssymb}
\usepackage{mathtools}
\usepackage{amsthm}

\usepackage[capitalize,noabbrev]{cleveref}

\theoremstyle{plain}
\newtheorem{theorem}{Theorem}[section]
\newtheorem{proposition}[theorem]{Proposition}
\newtheorem{lemma}[theorem]{Lemma}

\theoremstyle{definition}
\newtheorem{definition}[theorem]{Definition}

\theoremstyle{remark}
\newtheorem{remark}[theorem]{Remark}

\usepackage[textsize=tiny]{todonotes}

\usepackage{etoc}
\etocdepthtag.toc{mtchapter}
\etocsettagdepth{mtchapter}{subsection}
\etocsettagdepth{mtappendix}{none}
\usepackage[resetlabels]{multibib}
\usepackage{multirow}
\usepackage{multicol}
\usepackage{enumitem}
\usepackage[normalem]{ulem}

\usepackage{xcolor}
\usepackage{threeparttable}

\definecolor{cite_color}{HTML}{114083}
\definecolor{link_color}{RGB}{153, 0,0}  
\definecolor{url_color}{RGB}{153, 102,  0}
\definecolor{emp_color}{RGB}{0,0,255}
\hypersetup{
 colorlinks,
 citecolor=cite_color,
 linkcolor=link_color,
 urlcolor=url_color}
 
\usepackage{caption}
\usepackage{pifont}
\usepackage{makecell}
\usepackage{graphicx}
\usepackage{subfigure} 
\usepackage{amsmath}
\usepackage{multirow}
\usepackage{multicol}
\usepackage{enumitem}
\usepackage{ulem}
\usepackage{threeparttable}
\usepackage{wrapfig}
\usepackage{scrextend}
\usepackage{thm-restate}
\usepackage{array,hhline}
\usepackage{amsmath,amsfonts}
\usepackage{graphicx}
\usepackage{tabularborder}

\usepackage{framed} 
\definecolor{shadecolor}{rgb}{0.94, 0.97, 1.0}

\newcommand{\Eqref}[1]{Eq.~\eqref{#1}}
\newcommand{\mbf}[1]{\mathbf{#1}}

\newcommand{\mcal}[1]{\mathcal{#1}}

\def\Done{}

\def\W{\mathbf{W}}
\def\L{\mathbf{L}}
\usepackage{amssymb} 
\def\R{\mathbb{R}}
\def\I{\mathbf{I}}
\def\U{\mathbf{U}}
\def\u{\mathbf{u}}
\def\Q{\mathbf{Q}}

\def\x{\mathbf{x}}

\def\X{\mathbf{X}}

\def\xx{\times}
\def\V{\mathcal{V}}
\def\E{\mathcal{E}}
\def\G{\mathcal{G}}

\def\H{\mbf{H}}
\def\D{\mbf{D}}
\def\W{\mbf{W}}
\def\T{\mbf{T}}
\def\P{\mbf{P}}
\def\L{\mbf{L}}
\def\K{\mbf{K}}

\icmltitlerunning{Hypergraph Convolutional Networks via Equivalence Between Hypergraphs and Undirected Graphs}

\begin{document}

\twocolumn[
\icmltitle{Hypergraph Convolutional Networks via Equivalence\\ Between Hypergraphs and Undirected Graphs}



\icmlsetsymbol{equal}{*}

\begin{icmlauthorlist}
\icmlauthor{Jiying Zhang}{equal,comp,yyy}
\icmlauthor{Fuyang Li}{equal,yyy}
\icmlauthor{Xi Xiao}{yyy}
\icmlauthor{Tingyang Xu}{comp}
\icmlauthor{Yu Rong}{comp}
\icmlauthor{Junzhou Huang}{sch}
\icmlauthor{Yatao Bian}{comp}
\end{icmlauthorlist}

\icmlaffiliation{comp}{Tencent AI Lab}
\icmlaffiliation{yyy}{Shenzhen International Graduate School, Tsinghua University}
\icmlaffiliation{sch}{University of Texas at Arlington}
\icmlcorrespondingauthor{Yatao Bian}{yataobian@tencent.com}
\icmlcorrespondingauthor{Xi Xiao}{xiaox@sz.tsinghua.edu.cn}

\icmlkeywords{Machine Learning, ICML}

\vskip 0.3in
]



\begin{abstract}
As a powerful tool for modeling  complex relationships, hypergraphs are gaining popularity from the graph learning community. However, commonly used frameworks in deep hypergraph learning focus on hypergraphs with \textit{edge-independent vertex weights}~(EIVWs), without considering hypergraphs with \textit{edge-dependent vertex weights} (EDVWs) that have more modeling power. 
To compensate for this, we present General Hypergraph Spectral Convolution~(GHSC), a general learning framework that not only  handles EDVW and EIVW hypergraphs, but more importantly, enables theoretically explicitly utilizing the existing powerful Graph Convolutional Neural Networks~(GCNNs) such that largely ease the design of Hypergraph Neural Networks.
In this framework, the graph Laplacian of the given undirected GCNNs is replaced with a unified hypergraph Laplacian that incorporates vertex weight information from a random walk perspective by equating our defined generalized hypergraphs with simple undirected graphs.
Extensive experiments from various domains including social network analysis, visual objective classification, and  protein learning demonstrate the state-of-the-art performance of the proposed framework. \footnote{This work is done when Fuyang Li  works as an intern in Tencent AI Lab.}
\end{abstract}

\printAffiliationsAndNotice{\icmlEqualContribution} 


\section{Introduction}

Hypergraphs, whose edges link to the arbitrary number of vertices, attract much attention recently from researchers in the graph learning community~\citep{hgnn,DHGNN,UniGNN,zhang2022learnable}. Hypergraph is capable of modeling high-order complex relationships that a simple graph fails to capture, making it a great potential for solving real-world problems~\cite{chien2021allset,benson2016higher,benson2017spacey}.

The topology of the hypergraph is embedded in vertex weights,
(a generalized incident matrix), 
which can be roughly divided into two categories: edge-independent and edge-dependent~\citep{2019random,hayashi2020hypergraph}, depending on whether the vertex weights are related to the incident hyperedges or not~(EDVW-hypergraph implies that each vertex $v$ is assigned a weight $q_e(v)\in\R$ for each incident hyperedge $e$). Many hypergraph neural networks have been proposed ~\citep{hgnn,DHGNN,yadati2020neural,UniGNN} to handle EIVW-hypergraph. 
Yet, a rigorous investigation of EDVW-hypergraph is still lacking, even though they enjoy greater expressive power~\citep{ding2010interactive,huang2010image,li2018tail,zhang2018dynamic}.

There exists several heuristic message passing frameworks  for EIVW-hypergraph learning~\cite{UniGNN,chien2021allset}. However, their frameworks required  careful design and take no advantage of the existing graph learning algorithms. 
We thus aim in this paper to develop a general framework that can utilize existing GNNs to handle both EDVW and EIVW hypergraphs.
We are devoted to spectral-based methods, which are considered to be robust and allow for simple model property analysis in simple graph learning~\citep{wu2020comprehensive}.

From a random walk perspective, the focus in previous works \citep{2019random} has been on transforming the EIVW-hypergraph and EDVW-hypergraph into respective undigraphs and digraphs in order to handle hypergraphs. 
Based on it, one can directly use the digraph Neural Networks to learn hypergraphs via this well-established equivalence induced Laplacian. 
Despite this, most convolutional algorithms for digraphs are more or less related to undirected graphs
 \citep{monti2018motifnet,li2020scalable,ma2019spectral,tong2020directed}. Essentially, they transform digraphs to undigraphs with various techniques due to the difficulties of directed graph Laplacian. 
\begin{figure*}
    \centering
    \includegraphics[width=1\linewidth]{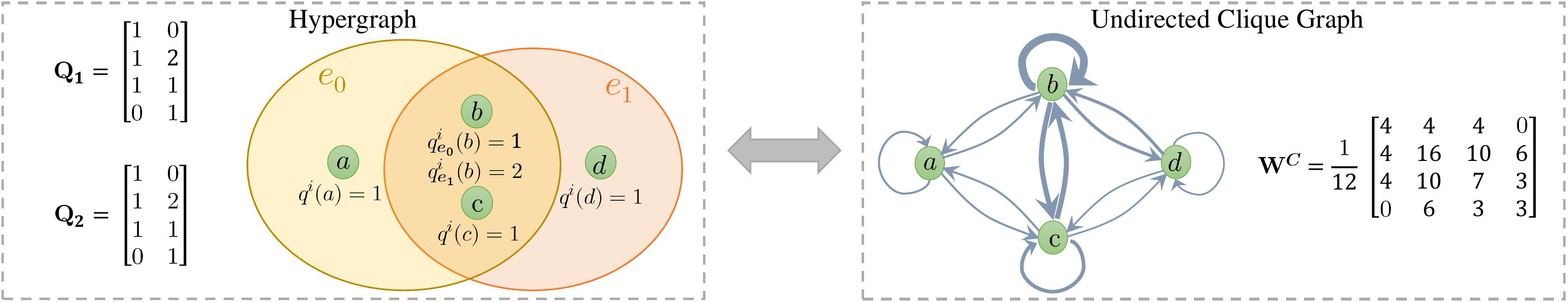}
     \caption{An example of hypergraph and its equivalent weighted undirected graph~($\W^{C}=(\W^{C})^{\top}$), where $q^i(\cdot)=Q_i(\cdot,e), i\in\{1,2\}$. Here, $\Q_1=\Q_2$ are both edge-dependent. The characteristics of the hypergraph are encoded in the weighted incidence matrices~(i.e. vertex weights) $\Q_1, \Q_2$.
    $\W^{C}:=\Q_2\D_e^{-1}\Q_2^{\top}$ denotes the edge-weight matrix of the clique graph  
     and can be viewed as the embedding of high-order relationships.}
    \label{fig:hypergraph2graph_connected}
\end{figure*}
An arguably  more succinct  way for hypergraph learning is to use undigraph convolutional networks via transforming hypergraphs to undigraphs.
Compared to digraphs, the undigraph convolutional networks have been extensively studied and there exist various effective strategies for analyzing their theoretical  properties, to name a few  \citep{gcnii,zhu2021simple,pairnorm,wu2020comprehensive}.
It remains unclear, however, whether EDVW-hypergraphs can be made equivalent to undigraphs in a principled manner, which is a major obstacle to this promising prospect.

In this paper, we investigate the equivalence condition between hypergraphs and undigraphs (Thm.\ref{Th:EquivalencyGrpah2Hyeprgrpah}) by defining a unified random walk and a generalized hypergraph that contains vertex weights. 
Then a unified Laplacian that embeds vertex weights is derived from the equivalence condition and is applied to replace the graph Laplacian of existing GCNNs for constructing the GHSC framework. 
Therefore, GHSC framework is capable of handling both EDVWs and EIVWs. 
Experimental results on EIVW and EDVW hypergraphs indicate that our framework achieves the best performance among all baselines, demonstrating the effectiveness and generalizability of our methodology.

\textbf{Contributions}:
1) We define a generalized hypergraph capturing EDVWs via designing a two-step unified random walk. Moreover, we construct the equivalency conditions between the generalized hypergraph and corresponding undigraph, leading to a unified Laplacian that embeds the information of EDVWs.
2) We propose a General Hypergraph Spectral Convolution framework named GHSC,  that can utilize existing graph neural  networks  for both EIVW and EDVW hypergraph learning via replacing the graph Laplacian with the unified Laplacian.
3) Extensive experiments across various domains (social network analysis, visual objective, and protein classification ) demonstrate the generalization and effectiveness of the proposed methods on EIVW-hypergraph and EDVW-hypergraph learning tasks. Notably, we are the first to adopt EDVW-hypergraphs for protein structure modeling and achieve a significant performance boost.

Due to the space limit, more discussions on related works, all proof and remarks are  deferred to Appendix.
\vspace{-2mm}

\begin{wrapfigure}{R}{0.60\linewidth}
    \centering
    \includegraphics[width=0.95\linewidth]{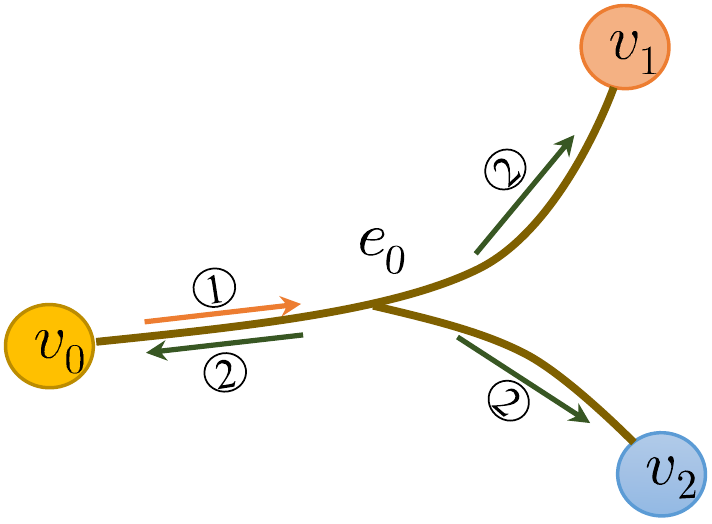}
    \caption{\small Hypergraph Random Walk. \ding{192} denotes the first step and \ding{193} represents the second.}
    \label{fig:random_walk}
\end{wrapfigure}

\section{The Theory of Equivalence }
\paragraph{Notations.}
$\I\in\R^{n\xx n}$ denotes the identity matrix,
and $\mbf 1\in \R^{n}$ represents the vector with ones in each component.
We use boldface letter $\x\in \R^n$
 to indicate an $n$-dimensional vector, where $x(i)$ is
the $i^\text{th}$ entry of $\x$. We use a boldface capital letter
$\mbf A\in\R^{m\times n}$ to denote an $m$ by $n$ matrix and use $A(i,j)$
to denote its ${ij}^\text{th}$ entry.

Let $\G(\V,\E,\omega)$ be a graph with vertex set $\V$, edge set $\E$, and edge weights $\omega$. 
Let $\mathcal{H}(\V,\E,\W,\Q)$ be the hypergraph with vertex set $\V$, edge set $\E$. Let $w(e)$ denote the weight of hyperedge $e$. $\W\in \R^{|\E|\xx|\E|}$ is the edge-weights diagonal matrix with entries $W(e,e)=w(e)$. $\Q$ denotes edge-vertex-weights matrix with entries $Q(u,e)=q_{e}(u)\in\R$ if $u\in e$ and 0 if $u\notin e$. $\Q$ is said to be \textit{edge-independent} if 
$q_{e}(u) = q(u)\in \R$ for all $e\ni u$, and called \textit{edge-dependent} otherwise~\citep{2019random}. If $q_{e}(u)$ equals to $1$ for all linked $u$ and $e$  
,$\Q$ would reduce to the binary incident matrix $\H$, which is widely used to represent the structure of hypergraph~\citep{zhou2006learning,carletti2021random}. The matrix $\Q$ can also be viewed as a weighted incident matrix.
Note that a graph is a special case of a hypergraph with hyperedge degree $|e|=2$. 
We say a hypergraph is connected when  its unweighted clique graph~\citep{2019random} is connected and we assume hypergraphs are connected. Throughout this paper, equivalence refers to equivalence between hypergraphs and  undigraphs if not otherwise specified.

\subsection{Unified Random Walk and Generalized Hypergraphs}
Traditionally, hypergraph random walk is defined by a \emph{two-step} manner~\citep{zhou2006learning,ducournau2014random}~
(Fig. \ref{fig:random_walk}).
Then,~\citet{2019random} raise a new random walk involving the edge-dependent vertex weights into the second step and build the equivalency between EDVW-hypergraphs and digraphs.
However, due to the limitation of their hypergraph definition,
they fail to answer whether the EDVW-hypergraphs can be equal to undigraphs. So in this part, we integrate the existing two-step random walk methods~\citep{2019random,carletti2020random,carletti2021random} to obtain a comprehensive~\textit{ unified random walk} with the vertex weights added into the first step, based on which we further define a generalized hypergraph.
\begin{definition}[Unified Random Walk on Hypergraphs]\label{def:random_walk_Definition}
     The unified random walk on a hypergraph $\mathcal{H}$ is defined in a 
     two-step manner: Given the current vertex $u$, 
     \vspace{-2mm}
     
    \noindent \textbf{Step I}:  choose an arbitrary hyperedge $e$ incident to $u$, with the probability
    \begin{equation}
    \small
        p_1 = \frac{w(e) \sum_{v\in \V} Q_2(v, e) \rho(\sum_{v\in \V} Q_2(v,e))Q_1(u,e)}{\sum_{e \in \E} w(e) \sum_{v\in \V} Q_2(v,e) \rho(\sum_{v\in \V} Q_2(v,e))Q_1(u,e)}; \label{eq:p1}
    \end{equation}
    \textbf{Step II}:  choose an arbitrary vertex $v$ from $e$, with the probability 
    \begin{equation} \label{eq:p2}
    p_2=\frac{Q_2(v,e)}{\sum_{v\in \V} Q_2(v,e)},    
    \end{equation}
    \vspace{-0.2mm}
\noindent where $w(e)$ denotes the hyperedge weight, $Q_1(u,e)$ denotes the contribution of hyperedge $e$ to vertex $u$  in the first step while $Q_2(v,e)$ represents the contribution of vertex $v$ to hyperedge $e$ in the second step. $\rho(\cdot)$ is a real-valued function that  acts on the degree of the hyperedge and is used to control the random process. We set $\rho = (\cdot)^{\sigma}$~(power function) for example. For positive values of $\sigma$, the hyperedges with larger degree will dominate the random process. Conversely, when $\sigma$ is negative, hyperedges with small degree are likely to drive the random walk process.
\end{definition}
Suppose $\delta(e):= \sum_{v\in \V} Q_2(v,e)$ is the degree of hyperedge, and $d(v) := \sum_{e \in \E} w(e)\delta(e)\rho(\delta(e))Q_1(v,e)$ is the degree of vertex $v$. The transition probability of our unified random walk on a hypergraph from vertex $u$ to vertex $v$ is:
  \begin{align}
  \small
    \label{Eq:tansition_propobility}
      P(u,v)
      =\sum_{e \in \E} \frac{w(e)Q_1(u,e)Q_2(v,e)\rho(\delta(e))} {d(u)}.
  \end{align}
Here, $P(u,v)$
can be written in a ${|\V|\xx|\V|}$ matrix form: $\P=\D_v^{-1}\Q_1 \W \rho(\D_{e})\Q_2^{\top}$, where $\D_v$, $\D_e$ are diagonal matrices with entries $D_v(v,v)=d(v)$ and $D_e(e,e)=\delta(e)$, respectively. $\rho(\D_e)$ represents the function $\rho$ acting on each element of $\D_e$. Notably, the $\rho$ is possible to disappear in $\P$~(Appendix~\ref{proof:pro:k-uniform}) . 

The two introduced vertex weights $\Q_1$ and $\Q_2$ bring the following benefits: i) Providing the theoretical basis to investigate the equivalency between EDVW-hypergraphs and undigraphs. Thm. \ref{Th:EquivalencyGrpah2Hyeprgrpah} suggests that the equivalence depends on the relationship between these two $\Q$.
ii) Modeling the fine-grained high-order information.
The two matrices can be constructed heuristically to model the more fine-grained high-order information associated with the first step and the second step of random walk.  $\mathbf Q_1\neq \mathbf{Q_2}$ means that the contribution of $v$ to edge $e$ in the in-edge process is different from the out-edge process. 
More detailed intuition of our purpose to design $P(u,v)$ can be found in Appendix~\ref{intuition of UHRW}.
Notably, the existing hypergraph random walk ~\citep{zhou2006learning,2019random,carletti2020random,carletti2021random} can be seen as special cases with specific $p_1$, $p_2$~(see Appendix~\ref{app:special_case_and Laplacian}). 

The definition of the random walk is lazy since it allows self-loops~($P(v,v)>0$). Based on the unified random walk, we define the generalized hypergraph as follows.
\begin{definition}[Generalized Hypergraph]\label{def:generalized hypergraph}
     A generalized hypergraph is a hypergraph associated with the unified random walk in Definition \ref{def:random_walk_Definition}, denoted as  $\mathcal{H}(\V,\E,\W,\Q_1,\Q_2)$. Here, $\Q_1$, $\Q_2$ are vertex weights matrices, each of which can be edge-independent or edge-dependent. 
\end{definition}
\vspace{-2mm}
The generalized hypergraph is capable of carrying EDVWs and EIVWs information flexibly due the two $\Q$ introduced, laying the foundation for building the equivalence theory.

\subsection{Equivalence Between Hypergraphs and Undigraphs}\label{sec:Equivalency}
\begin{wrapfigure}{R}{0.6\linewidth} 

   \centering
            \vspace{-4mm}
    \includegraphics[width=1\linewidth]{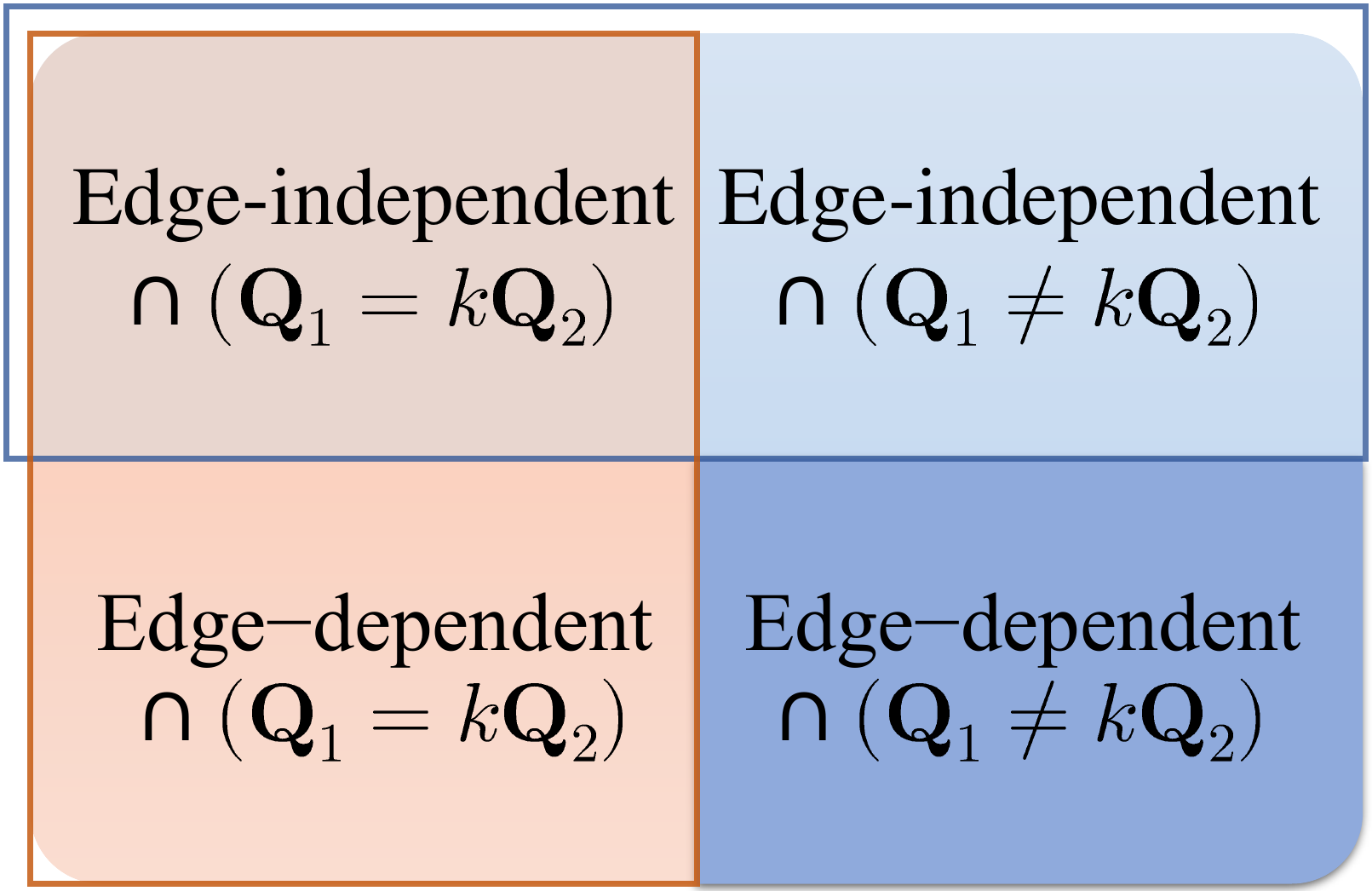}
    \caption{\small The division of Equivalency problem~(undigraph). The blue box indicates condition (1) in Thm. \ref{Th:EquivalencyGrpah2Hyeprgrpah} and red box indicates condition (2). The dark blue part is still open yet.}
    \label{fig:equvialencyPartition}
        \vspace{-4mm}
\end{wrapfigure}
In this part, 
we investigate the condition under which the generalized hypergraph is equivalent to a weighted undirected clique graph. This makes it possible to utilize undigraphs as low-order encoders of  hypergraphs thus enabling  hypergraph learning directly from the equivalent undirected graphs by means of  undigraph convolutional networks, that is,
any undigraph technique can be used as a subroutine for hypergraph learning. 

The clique graph of $\mathcal{H}(\V,\E,\W,\Q_1,\Q_2)$ is denoted as $\G^{C}$, which is an unweighted graph with vertices $\V$ and  edge set $\E_{\G^C} = \{(u,v):u,v\in e,e\in\E\}$. Actually, $\G^{C}$ turns all hyperedges into cliques.
We first present the definition of the equivalency in connection with Markov chains.
\begin{definition}
    \label{def:equi}
    (\citet{2019random}) Let $M_1,M_2$ be the Markov chains with the same finite state space, and let $\P^{M_1}$ and $\P^{M_2}$ be their probability transition matrices, respectively. $M_1,M_2$ are equivalent if  $\P^{M_1}_{u,v}=\P^{M_2}_{u,v}$  for all states $u$ and $v$. 
\end{definition}
It is easy to understand that any simple undigraph can be constructed as a special generalized
hypergraph, while the converse is not necessarily true. We provide a counterexample in Appendix \ref{app:Nonequivalent_condition}. This result implies that undigraph convolutional networks can not trivially be extended to generalized hypergraphs. 
So, we give one of the implicit equations for formulating the equivalency problem to explore the explicit equivalency conditions in Appendix~\ref{proof:le:EquivalencyUndigrpah2Hyeprgrpah}.
Based on it, we derive  the explicit condition under which random walks on hypergraphs are equivalent to undigraphs.
\begin{snugshade}
\vspace{-.2cm}
\begin{restatable}[Equivalency between generalized hypergraph and weighted undigraph]{theorem}{restathmtwo}
    \label{Th:EquivalencyGrpah2Hyeprgrpah}
    Let $\mathcal{H}(\V,\E,\W,\Q_1,\Q_2)$ denote the generalized hypergraph in Definition \ref{def:generalized hypergraph}.
    When $\mcal H$ satisfies any of the condition bellow:
    
    \emph{Condition (1):} $\Q_1$ and $\Q_2$ are both edge-independent;
     \\ \emph{Condition (2):} $\Q_1=k\Q_2 \;  (k\in\R)$,
     
     there exists a weighted undirected clique graph $\G^{C}$ such that a random walk on $\mathcal{H}$ is equivalent to a random walk on $\G^{C}$.
\end{restatable}
\vspace{-.1cm}
\end{snugshade}
\vspace{-.3cm}
It is easy to verify that  conditions (1) and (2) satisfy \Eqref{Eq:equivalencyEquation}.
This theorem brings insights from three aspects:
i) \textbf{Hypergraph foundation}. The equivalency itself is a fundamental problem and remains open before this work~\citep{agarwal2006higher,2019random}. Our conclusion provides more adaptive and explores more essential conditions thanks  to the unity of our random walk and the generalization of hypergraph defined~(Figure \ref{fig:equvialencyPartition}).  Notably,~\citet{2019random} shows that the equivalency condition is: $\Q_1=\H$ and $\Q_2$ is edge-independent (a special case of the condition~(1)), while Thm.   \ref{Th:EquivalencyGrpah2Hyeprgrpah} implies the equivalency is not only related to the dependent relationships between vertex weights and edges, but also to whether the vertex weights used in the first and second step of the unified random walk are proportional.
Furthermore, thanks to the introduced $\Q_1$ in the generalized hypergraph, we can obtain the equivalency between EDVW-hypergraph and undigraph~(i.e. condition (2) ), filling an important gap in the equivalency theory of EDVW-hypergraph.
ii) \textbf{Providing a theoretical basis for hypergraph applications.}
The condition~(2), containing both the edge-independent vertex weights and edge-dependent vertex weights cases,  which matches different hypergraph applications~\citep{ding2010interactive,zeng2016learn,zhang2018dynamic}. This theorem provides these applications with theoretical explanations. The condition~(2) also reveals that the existing random walks \citep{carletti2020random,carletti2021random} on hypergraphs are equivalent to undigraph.
iii) \textbf{Hypergraph learning. }
One can obtain hypergraph representations directly by exploiting the undigraph learning methods.
Especially, the condition (2), which implies that an EDVW-hypergraph can be equivalent to an undigraph, gives the theoretical guarantee for  learning hypergraphs without losing edge-dependent vertex weights via undigraph neural networks.

\subsection{Unified Hypergraph Laplacian} \label{sec:laplacian}
\vspace{-2mm}
To build a bridge between spectral convolution and equivalency, we derive the equivalent undirected weighted graph Laplacian based on Lemma \ref{le:EquivalencyUndigrpah2Hyeprgrpah} and Thm.~\ref{Th:EquivalencyGrpah2Hyeprgrpah}. Then the graph Laplacian can be considered as the  Laplacian of generalized hypergraph. 
Starting from the deduced Laplacian, it is direct and convenient to construct the spectral convolution for hypergraphs.
Formally, we have
\begin{table*}[t]
\def\p{$\pm$} 
\centering
\vspace{-2mm}
\setlength\tabcolsep{16 pt}
\caption{Summary of classificaiton accuracy(\%) results. We report the average test accuracy and its standard deviation over 10 train-test splits. (OOM: our of memory)}
    \vspace{-3mm}
\scalebox{0.89}{
    \begin{tabular}{l|l|c|c|c|c|c}
        \toprule 
              \multicolumn{1}{c}{Dataset}& \multicolumn{1}{c}{Architecture} &\multicolumn{1}{c}{\begin{tabular}[c]{@{}c@{}}Cora\\ (co-authorship)\end{tabular}}  & \multicolumn{1}{c}{\begin{tabular}[c]{@{}c@{}}DBLP\\ (co-authorship)\end{tabular}} & \multicolumn{1}{c}{\begin{tabular}[c]{@{}c@{}}Cora\\ (co-citation)\end{tabular}} & \multicolumn{1}{c}{\begin{tabular}[c]{@{}c@{}}Pubmed\\ (co-citation)\end{tabular}} & \multicolumn{1}{c}{\begin{tabular}[c]{@{}c@{}}Citeseer\\ (co-citation)\end{tabular}} \\
        \midrule
             MLP+HLR & - & 59.8\p4.7 & 63.6\p4.7 & 61.0\p4.1 & 64.7\p3.1 & 56.1\p2.6 \\
             FastHyperGCN & spectral-based & 61.1\p8.2 & 68.1\p9.6 & 61.3\p10.3 & 65.7\p11.1 & 56.2\p8.1 \\
             HyperGCN\Done & spectral-based & 63.9\p7.3 & 70.9\p8.3 & 62.5\p9.7& 68.3\p9.5 & 57.3\p7.3 \\
            HGNN & spectral-based & 63.2\p3.1 & 68.1\p9.6 & 70.9\p2.9& 66.8\p3.7 & 56.7\p3.8 \\
             HNHN & message-passing & 64.0\p 2.4 & 84.4\p 0.3& 41.6\p 3.1 & 41.9\p4.7 & 33.6\p 2.1  \\
             HGAT & message-passing & 65.4$\pm$1.5 & OOM &52.2$\pm$3.5 & 46.3$\pm$0.5 & 38.3$\pm$1.5 \\
             HyperSAGE & message-passing &  72.4\p1.6 & 77.4\p3.8 & 69.3\p2.7 &72.9\p1.3 & 61.8\p2.3 \\
             UniGNN &message-passing & 75.3\p1.2 & 88.8\p0.2 & 70.1\p1.4 & 74.4\p 1.0& 63.6\p 1.3 \\
        
        \midrule
            H-ChebNet &spectral-based  & 70.6$\pm$2.1 & 87.9$\pm$0.24 &  69.7$\pm$2.0 & 74.3$\pm$1.5 & 63.5$\pm$1.3 \\
            H-APPNP & spectral-based & \textbf{76.4$\pm$0.8} & \underline{89.4$\pm$0.18} & \underline{70.9$\pm$0.7} &  75.3$\pm$1.1 & \underline{64.5$\pm$1.4} \\
            H-SSGC & spectral-based & 72.0$\pm$1.2 & 88.6$\pm$0.16 &68.8$\pm$2.1  & 74.5$\pm$1.3 & 60.5$\pm$1.7  \\
            H-GCN & spectral-based & 74.8\p0.9 & 89.0\p 0.19 &{ 69.5\p 2.0} & \underline{ 75.4\p1.2} & 62.7\p1.2 \\
            H-GCNII & spectral-based & \underline{76.2$\pm$1.0} & \textbf{89.8$\pm$0.20} & \textbf{72.5$\pm$1.2}
            & \textbf{75.8$\pm$1.1} & \textbf{64.5$\pm$1.0} \\
        \bottomrule
    \end{tabular}
    }
    \vspace{-2mm}
    \label{tab:sota_accuray}    
\end{table*}

\begin{restatable}[]{corollary}{restathmthree}
    \label{Th:stationary_distribution}
  Let $\mathcal{H}(\V,\E,\W,\Q_1,\Q_2)$ be the generalized hypergraph in Definition \ref{def:generalized hypergraph}. Let $\hat{\D}_v$ be a $|\V|\xx|\V|$ diagonal matrix with entries $\hat{D}_v(v,v):=\hat{d}({v}) := \sum_{e\in \E}w(e)\delta(e)\rho(\delta(e))Q_2(v,e)$. No matter $\mcal H$ satisfies condition~(1) or condition~(2) in Thm.~\ref{Th:EquivalencyGrpah2Hyeprgrpah}, it obtains the unified explicit form of stationary distribution $\pi$ and Laplacian matrix $\L$ as:
\begin{align}
\small
    \pi = \frac{\mbf{1}^{\top}\hat{\D}_{v}}{\mbf{1}^{\top}\hat{\D}_{v}\mbf{1}} ~ \text{and} ~ 
    \L = \I-\hat{\D}_{v}^{-1/2}\Q_2\mathbf{W}\rho(\D_e)\Q_2^{\top}\hat{\D}_{v}^{-1/2}.   
    \label{Eq:transition_matrix}
\end{align}  

\end{restatable}
\vspace{-4mm}

 From  Corollary~\ref{Th:stationary_distribution}, $\L$ can be viewed as a normalized Laplacian led by the equivalent undigraph $\G^{C}$ with adjacency  matrix $\Q_2\W\rho(\D_e)\Q_2^{\top}$, based on which we can design the EDVW-hypergraph spectral convolutions~(The $\Q_1$ is indeed reduced in the derivation of the transition matrix).
Meanwhile, Corollary~\ref{Th:stationary_distribution} implies that when $\Q_1$, $\Q_2$ are both edge-independent or $\Q_1=k\Q_2$, we can directly use the Laplacian matrix $\L$ to analyze the spectral properties.  The spectral properties of Laplacian~\citep{chung1997spectral} are vital in the research of graph neural networks to design a stable and effective convolution operators. 
Therefore, we deduce two  
important conclusions concerning eigenvalues of $\L$ as the basic theory of Laplacian application under the equivalency conditions in Thm. \ref{Th:EquivalencyGrpah2Hyeprgrpah}. One claims the eigenvalues range of $\L$ is $[0,2]$ (details in Appendix \ref{proof:th:eignvectors}), and 
the other describes the rate of convergence of our unified random walk~(see Appendix \ref{proof:Th:convergency_rate}), which is important to analyse the property of our GHSC framework in section \ref{sec:convolution}~(details in Appendix \ref{proof:over_smoothing}).

\paragraph{Equivalence from a Transform View.} The equivalence can be viewed as transforming hypergraphs to undigraphs. Noted that condition(1) and (2) both lead to the same graph $\G^C$ with weights $\Q_2\W\rho(\D_e)\Q_2^{\top}$, indicating the transform is not an injection, that is, different hypergraphs are possibly mapped to the same graphs.
However, this does not prevent the transform from being a powerful conversion tool suitable for downstream tasks thanks to the elaboration of unified random walks. Indeed, one can design $\Q_1=\Q_2$ in practical for alleviating the information loss caused by the transform.  In our experiments, we all follow the setting $\Q_1=\Q_2$, under which the conversion from hypergraph $\mcal{H}(\mcal{V,E},\mbf{W},\Q_1,\Q_1)$ to undirected graph $\G^{C}$ is injective.

\section{Hypergraph Spectral Convolutions Derived from Undirected GCNNs
  }\label{sec:convolution}
 Intuitively, there are two possible routes for designing hypergraph deep learning framework: 1. message-passing; 2. spectral-based methods.
Message passing has proved to be a powerful tool for extracting information from hypergraphs or graphs.
However, on the one hand, researchers usually adopt heuristic ideas to design message-passing models, leading to a lack of
theoretical guarantees~\citep{UniGNN,chien2021allset}. This makes it difficult to analyze properties of corresponding neural networks
directly, such as over-smoothing issue~\citep{li2018deeper}. On the other hand, spectral-based
methods not only have a solid foundation in graph signal processing~\citep{kipf2016semi,hgnn}, 
but also effectively inspires the design of message passing techniques~\citep{xu2018how}.
 Here, we develop a hypergraph spectral convolutions framework, based on equivalency conditions in Thm.~\ref{Th:EquivalencyGrpah2Hyeprgrpah}. As the Laplacian $\L$ enjoys the same formula under any of the two equivalency conditions, algorithms deduced by $\L$ would be adaptive to any generalized hypergraph satisfying Thm.~\ref{Th:EquivalencyGrpah2Hyeprgrpah}, including EIVW-hypergraph and EDVW-hypergraph.

\paragraph{General Hypergraph Spectral Convolution Framework.}
The General  Hypergraph Spectral Convolutions~(GHSC) is defined as a general end-to-end framework that can utilize any GCNNs as a backbone,
\begin{align}
\label{eq:h-gnns}
    Y = g_f(\X,\L) = f(\X ,\L_g \gets  \L),
\end{align}
where $\L_g$ is the graph Laplacian used in undigraph GCNNs $f$.  $\gets$ denotes replacing the graph Laplacian with hypergraph Laplacian $\L$ that defined in Corollary \ref{Th:stationary_distribution}. $Y$ denotes the output of $g_f$.
We call these GNNs-induced Hypergraph neural network $g_f$ \textbf{H-GNNs}. For example,  H-GCN and H-SSGC represent hypergraph NNs induced by the GNN models GCN~\cite{kipf2016semi} and SSGC~\cite{zhu2021simple} , respectively. 
\begin{table*}[thbp]
\def\p{$\pm$} 
\centering
\setlength\tabcolsep{5pt} 
\caption{Test accuracy on visual object classification. Each model we ran 10 
random seeds and report the mean \p~standard deviation. BOTH means GVCNN+MVCNN, which represents combining the features or structures to generate multi-modal data. }
    \vspace{-2mm}
\scalebox{0.8}{
\begin{tabular}{c|cc|ccc|ccccc}
    \toprule 
    \multirow{1}{*} { Datasets }& \multirow{1}{*} { Feature } & Structure  & HGNN & UniGNN & HGAT & H-ChebNet&  H-SSGC & H-APPNP & H-GCN & H-GCNII \\
    \hline
    \multirow{3}{*}{\begin{tabular}[c]{@{}l@{}}NTU\end{tabular}} 
    & MVCNN &  MVCNN & 80.11\p0.38  & 75.25$\pm$0.17 & 80.40$\pm$0.47 & 78.04$\pm$0.46 & 81.23$\pm$0.24 & 80.16$\pm$0.36& \underline{81.37\p0.63}  &  \textbf{82.01$\pm$0.39}  \\
    & GVCNN &  GVCNN  & 84.26\p0.30  & 84.63$\pm$0.21 & 84.45$\pm$0.12 & 83.51$\pm$0.40 & 84.26$\pm$0.12 & \underline{84.96$\pm$0.41} & \textbf{85.15\p0.34 } &  84.34$\pm$0.54  \\
    & BOTH &  BOTH &  83.54\p0.50  & 84.45$\pm$0.40  & 84.05$\pm$0.36 & 83.16$\pm$0.46   & 84.13$\pm$0.34 &  83.57$\pm$0.42 & \underline{84.45\p0.40}  &   \textbf{85.17$\pm$0.36}  \\
    \hline
    \multirow{3}{*}{\begin{tabular}[c]{@{}l@{}}Model-\\Net40\end{tabular}} 
    & MVCNN       &  MVCNN       & 91.28\p 0.11  & 90.36$\pm$0.10 & 91.29$\pm$0.15 & 90.86$\pm$0.29 & 91.21$\pm$0.11 & 91.18$\pm$0.21  &  \underline{91.99\p0.16}            &  \textbf{92.03$\pm$0.22}  \\
    & GVCNN       &  GVCNN       &  92.53\p0.06  & \textbf{92.88$\pm$0.10} & 92.44$\pm$0.11 & 92.46$\pm$0.15 &  92.74$\pm$0.04 & 92.46$\pm$0.09 & 92.66\p 0.10  & 
    \underline{92.76$\pm$0.06}    \\
    & BOTH& BOTH  &   {97.15\p0.14}  & 96.69$\pm$0.07 & 96.44$\pm$0.15 & 96.95$\pm$0.09  & 97.07$\pm$0.07 &  97.20$\pm$0.14& \underline{97.28\p0.15}   &  \textbf{97.75$\pm$0.07}    \\
    \bottomrule
\end{tabular}
}
    \vspace{-2mm}
    \label{tab:object_accuracy} 
\end{table*}

The GHSC holds the following advantages: (1) It can handle both EIVW-hypergraphs and EDVW-hypergraphs via the unified Laplacian $\L$. (2) The unity of random walk makes it possible to aggregate more fine-grained information. (3) By using the framework, one can take established powerful techniques on undigraphs as a subroutine for hypergraph learning, which would largely ease hypergraph learning in real-world scenarios. (4) The GHSC is a spectral framework that may share the same properties as GCNNs, such as over-smoothing issues~\citep{li2018deeper}. Specifically, Corollary \ref{Th:convergency_rate} implies that the spectral convolution deduced from the Laplacian $\L$ might suffer from the over-smoothing issues~\citep{chen2022preventing}~(Appendix \ref{proof:over_smoothing}). This means the existing solutions of the over-smoothing issue also can be extend for hypergraph learning. 

\section{Empirical Studies}\label{sec:Experment}
We evaluate our proposed methods on two semi-supervised node classification tasks: citation network classification, visual object classification, and a graph classification task: fold classification. For simplicity, we set $\rho(\cdot)$ to be a power function $(\cdot)^{\sigma}$ ($\sigma$ is a hyper-parameter). 
 The weight matrix  of hyperedges $\W$, we set, to be an identity matrix by default. 
Citation network classification belongs to EIVW-hypergraph learning tasks~($\Q_1=\Q_2=\H$), and visual object classification and protein learning are EDVW-hypergraph learning tasks. Besides, in our experiments, we all follow the setting $\Q_1=\Q_2$, under which the conversion from hypergraph $\mcal{H}(\mcal{V,E},\mbf{W},\Q_1,\Q_1)$ to undirected graph is injective.
The details regarding experiments, Ablation Analysis, Running Time and Computational Complexity etc. can be found in Appendix \ref{sec:Details_of_Experiments}.

\vspace{-2mm}
\subsection{Citation Network Classification} \label{sub:citation_classification}
\vspace{-2mm}
The datasets we use are hypergraph benchmarks constructed by \citet{hypergcn}(See Appendix Table~\ref{tab:dataset_hypergcn}). We adopt the same public datasets~{(\url{https://github.com/malllabiisc/HyperGCN})} and train-test splits of \citet{hypergcn}. Note these datasets are EIVW-hypergraphs (i.e. $\Q_1=\Q_2=\H$). 
For baselines, we include Multi-Layer Perceptron with explicit Hypergraph Laplacian Regularization (MLP+HLR), HNHN~\citep{dong2020hnhn}~, HyperSAGE~\cite{arya2020hypersage}, HGAT~\citep{HyperGAT}, UniGNN~\citep{UniGNN}, and two recent spectral-based hypergraph convolutional neural networks~(HGCNNs): HGNN~\citep{hgnn} and HyperGCN~\citep{hypergcn}. 
To verify the effectiveness of the GHSC framework, we construct five H-GNNs variants: H-ChebNet, H-GCN, H-APPNP, H-SSGC, H-GCNII with  ChebNet~\cite{defferrard2016convolutional}, GCN~\cite{kipf2016semi},  APPNP~\cite{klicpera2018predict},  SSGC~\cite{zhu2021simple} GCNII~
\cite{gcnii} as $f$ in Eq. \eqref{eq:h-gnns}, respectively. 

\textbf{Comparison with SOTAs.} 
As shown in Table \ref{tab:sota_accuray}, the results successfully verify the effectiveness of our models and achieve a new SOTA performance across all five datasets. 
Observation (1): H-GNNs gain consistently improvement than the hypergraph-tailored baselines. This significant performance is  benefited from the GHSC framework that is capable of taking advantage of the existing powerful undigraph NNs.
Observation (2): HGNN, HNHN and HGAT show poor performance on disconnected datasets(e.g. Citeseer), mainly due to the values of the row corresponding to an isolated vertex in the adjacency matrix of equivalent undigraph are 0, leading direct loss of its vertex information~(an example can see Fig. \ref{fig:hypergraph2graph} in appendix). And H-GNNs utilize the re-normalization trick, which can maintain the features of isolated vertices during aggregation.

\begin{table}[t]
\def\p{$\pm$} 
\centering
\setlength\tabcolsep{2pt} 
\vspace{-0mm}
\caption{Comparison of our methods to others on  fold classification. We report the \textit{mean accuracy} (\%) of all proteins. }
    \vspace{-3mm}
\scalebox{0.78}{
 \begin{threeparttable}[b]
    \begin{tabular}{lcc|ccc}
        \toprule 
        \multirow{1}{*} { Methods } & \multirow{1}{*} { Architecture } & \multirow{1}{*} { \#params }  & 
        Fold & Super. & Fam. \\
        \hline
        
        \cite{hou2018deepsf} &  1D ResNet  &  41.7M & 17.0 & 31.0 &  77.0  \\
        \citet{rao2019evaluating}$^*$ &  1D Transformer  &  38.4M & 21.0 & 34.0 &  88.0  \\
        \citet{bepler2018learning}$^*$ &  LSTM  &  31.7M & 17.0 & 20.0 &  79.0  \\
        \citet{strodthoff2020udsmprot}$^*$ &  LSTM  &  22.7M    &  14.9 &21.5 &  83.6  \\
        \citet{kipf2016semi} &  GCNN   &  1.0M & 16.8 & 21.3  &  82.8  \\
        \citet{diehl2019edge}  &GCNN &1.0 M & 12.9 & 16.3 &72.5 \\
         \citet{gligorijevic2020structure}$^*$ & LSTM+GCNN &6.2M & 15.3 &20.6 & 73.2 \\
        \citet{baldassarre2020graphqa} & GCNN & 1.3M & 23.7 & 32.5  & 84.4 \\
        \hline
        HGNN &  HGCNN  &    2.1M &   24.2 &34.4 & 90.0  \\
        H-SSGC~(Ours) &   HGCNN  &  0.8M   &  21.4 & 23.0 &  78.4
        \\
        H-GCN~(Ours) &  HGCNN  &  2.1M  & {25.0} & { 36.3} & {91.6}  \\
        H-GCNII~(Ours) & HGCNN  & 0.6M  & \textbf{27.8} & \textbf{38.0} & \textbf{92.7} \\
        \bottomrule
    \end{tabular}

    \begin{tablenotes}
    \footnotesize
     \item[$^*$] Pre-trained unsupervised on 10-31 million protein sequences.
    \end{tablenotes}
\end{threeparttable}
}
\vspace{-6mm}
    \label{tab:Protein_fold}    
\end{table}

\vspace{-2mm}
\subsection{Visual Object Classification}\label{sub:object_classification}
\vspace{-2mm}
We employ two public benchmarks: Princeton ModelNet40
dataset~\citep{wu20153d} and the National Taiwan University~(NTU) 3D model dataset~\citep{chen2003visual} to evaluate our methods. We follow HGNN~\citep{hgnn} to preprocess the data by MVCNN \citep{su2015multi} and GVCNN~\citep{feng2018gvcnn} and obtain the EDVW-hypergraphs.  Finally, we use the datasets provided by its public Code~{(\url{https://github.com/iMoonLab/HGNN})}.

\textbf{Results.} 
From the results in Table \ref{tab:object_accuracy}, we can see our methods achieves competitive performance on both single modality and multi-modality~(BOTH) tasks and H-GCNII outperform baselines on multi-modality. These results reveal that our methods  have the advantage of combining such multi-modal information through concatenating the  weighted incidence matrices~($\Q$) of hypergraphs, which means merging the multi-level hyperedges. 

\vspace{-2mm}
\subsection{Protein Fold Classification}\label{sub:protein}
\vspace{-2mm}
We'd like to investigate whether hypergraphs are better than simple graphs for protein modeling, and the following results  confirm this conjecture.  The Motivation for constructing Protein Hypergraphs we deferred to Appendix \ref{subsec:QA_Fold}.
\vspace{-1mm}

\textbf{Protein hypergraph.}
A protein is a chain of amino acids (residues) that will fold to a 3D structure.
To simultaneously model protein sequence and spatial structure information, we build sequence hyperedges and distance hyperedges:  we choose $\tau$ consecutive amino acids $(v_i,v_{i+1},\cdots,v_{i+\tau})$ to form a sequence hyperedge, and choose amino acids whose spatial Euclidean distance is less than the threshold $\epsilon>0$  to form a spatial hyperedge, where $v_i(i=1,\cdots,|S|)$ represents the $i$-th amino acids in the sequence. Appendix \ref{subsec:QA_Fold} contains more details.

\textbf{Settings \& Results.} We use the well-known SCOPe 1.75 dataset~\citep{hou2018deepsf}
 and the cross-entropy loss.  
Table \ref{tab:Protein_fold} depicts that  H-GCNII outperform all others. It is manifest in the table that HGCNN-based methods achieve much better performance compared to GCNN-based. 
 The results demonstrated that the hypergraph can better model the 3D structure of proteins than the simple graph, and our proposed methods can learn a better protein representation. 
\vspace{-1mm}



\vspace{-1mm}
\section{Discussions and Future Work}
Despite the good results, there are some limitations which worth further explorations in the future: 1) We just provide sufficient conditions for the equivalency to undigraphs, but the necessary condition is still unclear.
2) With the equivalency between hypergraphs and undigraphs, we have extended the undigraph-based GCNNs to hypergraph learning in this paper.   Meanwhile,  the equivalency between hypergraphs and digraphs would allow one to extend digraph-based GCNNs to potentially  learn richer information in  hypergraphs.
3) It is worthwhile to incorporate  advances in GNN architectures, such as the $^p$GNNs
\citep{fu2021p}, to further improve the performance on hypergraphs. 
4) One could use the proposed unified random walk on hypergraphs  to devise new clustering algorithms for hypergraph partitioning.

\section*{Acknowledgements}
This work was supported in part by the National Natural Science Foundation of China (61972219), the Research and Development Program of Shenzhen (JCYJ20190813174403598, SGDX20190918101201696), the National Key Research and Development Program of China (2018YFB1800204), the Overseas Research Cooperation Fund of Tsinghua Shenzhen International Graduate School (HW2021013) and Shenzhen Science and Technology Innovation Commission (Research Center for Computer Network (Shenzhen Ministry of Education).

\bibliography{0_main}
\bibliographystyle{icml2022}

\newpage
\onecolumn
\appendix
\begin{center}
\Large
\textbf{Appendix}
 \\[20pt]
\end{center}

\etocdepthtag.toc{mtappendix}
\etocsettagdepth{mtchapter}{none}
\etocsettagdepth{mtappendix}{subsection}

{\small \tableofcontents}

\section{Related Work}
\label{appendix_related_work}

\subsection{\textcolor{blue}{Simplified Related Work}}
 \textbf{Deep learning for hypergraphs.}
Many hypergraph deep learning algorithms can essentially be viewed as redefining hypergraph propagation schemes on equivalent clique graph~\cite{agarwal2006higher} or its variants~\cite{hgnn,hypergcn,dong2020hnhn}. On the other hand, \cite{UniGNN} and \cite{chien2021allset} propose a message-passing and set function learning framework that covers the MPNN~\cite{MPNN} paradigm.
However, these frameworks fail to provide guarantees. Furthermore, they depend on the elaborate design of propagation schemes.
It remains open in deep hypergraph learning are: (i) a framework that can handle EDVW-hypergraphs (ii) a general theoretical framework that can directly utilize the  build on the Laplacian of hypergraph.

\textbf{Graphs Neural Networks.}
GNNs have received significant attention recently and many successful architectures have been raised for various graph learning tasks~\cite{kipf2016semi,MPNN,velivckovic2018graph, chen2021diversified, zhang2022fine}. Spectral-based GNNs have a solid foundation in graph
signal processing~\cite{shuman2013emerging} and allow for simple model property analysis. There exist many powerful spectral-based models such as ChebNet~\cite{defferrard2016convolutional}, GCN~\cite{GCN},SSGC~\cite{zhu2021simple}, GCNII~\cite{gcnii}, APPNP~\cite{klicpera2018predict}. Nevertheless, these methods only employ graphs and not hypergraphs. Fortunately, in our framework, these spectral methods can be extended  to hypergraphs.
 
\textbf{Equivalence between hypergraphs and graphs.} A fundamental problem in the spectral theoretical study of hypergraphs is the equivalency with undigraphs. So far, equivalence is established from two perspectives:the spectrum of Laplacian~\citep{agarwal2006higher,hayashi2020hypergraph} or random walk~\citep{2019random}. 
An important application of building up the equivalence between hypergraphs and graphs is to transform hypergraphs to clique graphs~\cite{zhou2006learning,agarwal2006higher,hein2013total,2019random}.

\textbf{Hypergraph with edge-dependent vertex weights.}
EDVW-hypergraphs have been widely adopted in  machine learning  applications, including 3D object classification~\citep{zhang2018dynamic}, image segmentation~\citep{ding2010interactive}, e-commerce~\citep{li2018tail}, image search~\citep{huang2010image}, hypergraph clustering~\citep{hayashi2020hypergraph} , text ranking~\cite{bellaachia2013random}, etc. However, the design of deep learning algorithms for processing EDVW-hypergraphs remains an open question.

\subsection{Detailed Related Work}

\paragraph{Deep learning for hypergraphs}
\citet{hgnn} introduce the first hypergraph architecture \textit{hypergraph neural network}(HGNN), based on the hypergraph Laplacian proposed by \citet{zhou2006learning}, however, 
a flaw in the theory makes the learning of unconnected networks very inefficient~(details can see Appendix \ref{app:hgnn}).  
\citet{hypergcn} use the non-linear Laplacian operators~\cite{chan2020generalizing} to convert hypergraphs to simple graphs by reducing a hyperedge to a subgraph with edge weights related only to its degree, which causes the information loss of hypergraphs and limited the generalization.
\citet{DHGNN} proposed a \textit{dynamic hypergraph neural networks} to update hypergraph structure during training. ~\citet{wendler2020powerset} utilize signal processing theory to define the convolutional neural network on set functions. \citet{zhang2020hypergraph} combined hypergraph label propagation with deep learning and introduced a \textit{Hypergraph Label Propagation Network} to optimize the hypegraph learning. \citet{yadati2020neural} proposed a \textit{generalised-MPNN} on hypergraph which unifies the existing MPNNs~\citep{MPNN} on simple graph and also raised a \textit{MPNN-Recursive} framework for recursively-structured data processing, but it performs poorly for other high-order relationships.
~\citet{UniGNN} generalized the current  graph message passing methods to hypergraphs and obtain a UniGCN with self-loop and a deepen hypergrpah network UniGCNII. However, it does not give reasons why self-loop can assist UniGCN to gain the improvement of performance in the citation network benchmark. Moreover, its experimental results with various depths may show that the HGCNNs have over-smoothing issue, but theoretical analysis to explain this phenomenon is lacked. Hypergraph deep learning has also been applied to many areas, such as ~NLP~\citep{HyperGAT}, computer vision~\citep{jin2019hypergraph}, recommendation system~\citep{sagnn}, etc
\paragraph{Hypergraph random walk and spectral theory.}
In machine learning applications, the Laplacian is an important tool to represent graphs or hypergraphs. While random walk-based graph Laplacian has a well-studied spectral theory, 
the spectral methods of hypergraphs are surprisingly lagged. The research of hypergraph Laplacian can probably be traced back to \citet{chung1993laplacian}, which defines the Laplacian of k-uniform hypergraph~(each hyperedge contains the same number of vertices).
Then ~\citet{zhou2006learning} defined a two-step random walk-based Laplacian for general hypergraphs.
Based on it, many works try to design a more comprehensive random walk on hypergraphs.  \citet{2019random} designed a random walk which considered edge-dependent vertex weights of hypergraphs in the \textit{second step} to replace the edge-independent weights in ~\citet{zhou2006learning}. Actually, the edge-dependent vertex weights could model the contribution of vertex $v$ to hyperedge $e$ and were widely used in machine learning  
such as 3D object classification~\citep{zhang2018dynamic,hgnn} and image segmentation~\citep{ding2010interactive} etc., but without spectral guarantees.

On the other hand, \citet{carletti2020random,carletti2021random} takes another perspective to gain more fine-grained information from a hypergraph by taking into account the degree of hyperedges to measure the importance between vertices in the \textit{first step}.
However, a comprehensive random walk should consider the above two aspects at the same time.
Furthermore,
different from the two-step random walk Laplacian of ~\citet{zhou2006learning},
\citet{lu2011high} and \citet{aksoy2020hypernetwork} define a $s$-th Laplacian through random $s$-walk on hypergraphs.
More recently, non-linear Laplacian, as an important tool to represent hypergraphs, has been extended to several settings, including directed hypergraphs~\citep{zhang2017re,chan2019diffusion}, submodular hypergraph~\citep{li2017inhomogeneous} etc.
Meanwhile, linear Laplacian has also been developed. For example, \citet{2019random}  use a two-step random walk to develop a spectral theory for hypergraphs with edge-dependent vertex weights, and
a two-step random walk on hypergraph  proposed by \citet{carletti2020random,carletti2021random} also designed a two-step random walk by involving the degree of hyperedges in the first step and lead a linear Laplacian.
In this work, we design a  unified two-step  random walk framework to study the linear Laplacians.

\paragraph{Equivalency between hypergraphs and undigraphs.}
A core problem in the spectral theoretical study of hypergraphs is the equivalency between hypergraphs and graphs. Once equipped with equivalency, it is possible to represent hypergraphs with lower-order graphs, thus reducing the difficulty of hypergraph learning. So far, equivalence can be established from two perspectives: spectrum of Laplacian or random walks. 
For the aspect of Laplacian spectrum, 
\citet{agarwal2006higher} firstly showed the  Laplacian of \citet{zhou2006learning} is equivalent to the Laplacian of the corresponding star expansion graph~(having the same eigenvalue problem). Next, \citet{hayashi2020hypergraph} claims that the Laplacian defined by~\citet{2019random} is equal to an undigraph Laplacian, but the undigraph Laplacian needs to be constructed by invoking the stationary distribution, which restricts its application. Meanwhile, from the random walk perspective, ~\citet{2019random} claims that hypergraph is equivalent to its clique graph~(undirected) when the vertex weight of hypergraph is edge-independent. However, despite the edge-dependent weights were widely used~\citep{ding2010interactive,zeng2016learn,zhang2018dynamic}, the equivalency problem remains open when
the vertex weight is edge-dependent.

\paragraph{Protein learning.}
Proteins are biological macromolecules with spatial structures formed by folding chains of amino acids, which have an important role in biology. A protein has a three-level structure: primary, secondary, and tertiary, where primary~(sequence of amino acids) and tertiary~(3D spatial) structures are often used to model proteins for representation learning. Based on amino acids sequence, there exist many related works of protein learning, such as \citet{rao2019evaluating,bepler2018learning,strodthoff2020udsmprot,AngularQA}, etc. Meanwhile,  many 3D structure based models are also raised for protein learning, such as \citet{VoroMQA,3DCNN,baldassarre2020graphqa,diehl2019edge, baldassarre2020graphqa,gligorijevic2020structure,hermosillaintrinsic}, etc. 
However, to the best of our knowledge, despite the hypergraphs have been developed to model proteins~\citep{maruyama2001learning}, there is still no related work to design the EDVW-hypergraph-based protein learning algorithm. 
 \section{Relations with Previous Random Walks on Hypergraph}\label{app:relation_to_previous}
 
 \subsection{A Classical Random walk on Hypergraph }
It can be traced back to \citet{zhou2006learning} in which they have given a view of random-walk to analyze the hypergraph normalized cut. The transition probability of \citet{zhou2006learning} from current vertex $u$ to next vertex $v$  is denoted as:
\begin{equation}
P(u, v)=\sum_{e \in \E} w(e) \frac{H(u,e)}{d(u)} \frac{H(v,e)}{\delta(e)}.
\end{equation}
It is easy to find $P(u,v)$ means: (i) choose an arbitrary hyperedge $e$ incident with $u$ with probability $w(e)/d(u)$ where $d(u) = \sum_{e \in \E} w(e)H(u,e)$; (ii) Then choose an arbitrary vertex $v \in e$ with probability  $1/\delta(e)$ where $\delta(e) = \sum_{v\in \V} H(v,e)$, which means to select vertex randomly in the hyperedge.

\subsection{Intuition and Analysis of The Unified Random Walk on Hypergraph}\label{intuition of UHRW}
In this subsection, we would like to explain our intuition of developing the unified random walk framework in Definition \ref{def:random_walk_Definition} from a view of two-step random walk on a hypergraph.
To generalize the notation of hypergraph random walk in the seminal paper~\cite{zhou2006learning}, in this work, we tend to explain the process from another perspective. The fact in~\citet{zhou2006learning} that 
\begin{align}
    d(u) &= \sum_{e\in \E}w(e)H(u,e) = \sum_{e \in \E} \sum_{v \in \V} w(e)\frac{H(u,e)H(v,e)}{\delta(e)}
    =\sum_{v \in \V}\sum_{e \in E(u,v)} \frac{w(e)}{\delta(e)}
\end{align}
and
\begin{align}
P(u, v)&=\sum_{e \in \E} \frac{ w(e)H(u,e)}{d(u)}\cdot \frac{H(v,e)}{\delta(e)}
     =  \frac{1}{d(u)}\sum_{e \in E(u,v)} \frac{w(e)}{\delta(e)}
     =\frac{\sum\limits_{e \in E(u,v)} \frac{w(e)}{\delta(e)}}{\sum\limits_{b \in \V}\sum\limits_{e\in E(u,b)} \frac{w(e)}{\delta(e)}}
\end{align}

show that the probability $P(u,v)$ also means a normalized weighted adjacency relationship between $u$ and $v$ (Here, $E(u,v)$ denotes the hyperedges contained vertices $u$ and $v$.). Furthermore, the relationship is actually measured by the sum of $\frac{w(e)}{\delta(e)}$ for all $e \in \E$ concluding pair $\{u,v\}$, which demonstrates that a hyperedge $e$ with larger degree~($\delta(e)$) linked to $\{u,v\}$ contribute less to the transition probability between $u$ and $v$.

From \citet{carletti2021random}, which proposed a random walk on hypergraphs~(a specital case of our unified random walk with $\Q_1=\Q_2=\H$ and $\rho(\cdot)=(\cdot)^\sigma$), we further explain the intuition of our purpose to design the $P(u,v)$ of our unified random walk. When $\Q_1=\Q_2=\H$, $P(u,v)$ in \Eqref{Eq:tansition_propobility} can be expressed as:
$$
P(u,v) = \frac{\sum\limits_{e\in E(u,v)}w(e)\rho(\delta(e))}{\sum\limits_{b\in \V}\sum\limits_{e\in E(u,b)}w(e)\rho(\delta(e))}
$$
That is to say, the influence of hyperedge degree should not be limited to an inverse relationship by introducing the function $\rho(\cdot)$.
Thus, by evolving more fine-grained vertexes weights $Q_{1}(u,e)$ and $Q_{2}(v,e)$ to replace $H(v,e)$ and $H(u,e)$, we obtain the final transition probability of our unified hypergraph random walk in \eqref{Eq:tansition_propobility}.

\subsection{Special Cases of The Unified Random Walk Framework and Laplacian}\label{app:special_case_and Laplacian}

We show the existing random walks and Laplacians are the special cases of our unified random walks and our unified Laplacians, respectively.
 Under our framework, the existing random walk models can be seen as a special case with specific $p_1$, $p_2$. It is easy to construct the relations of our framework to previous works as follows:
\begin{remark}[\citet{zhou2006learning}]
\label{Remark:zhou2006}
    When we choose $\rho(\cdot)=(\cdot)^{-1}$ and $\Q_1=\Q_2=\H$ in our framework, the probabilities of unified random walk on hypergraph are reformulated as $p_1=\frac{w(e)H(u,e)}{\sum_{e\in \E}w(e)H(u,e)}$ and $p_2=\frac{H(v,e)}{\sum_{v\in \V}H(v,e)}$. As the special case of ours satisfying $\Q_1=\Q_2=\H$ and $\rho(\cdot)=(\cdot)^{-1}$, \cite{zhou2006learning}'s random walk satisfies both condition~(1) and condition~(2) in Thm.~\ref{Th:EquivalencyGrpah2Hyeprgrpah}. So we obtain from Corollary~\ref{Th:stationary_distribution} that $\pi_{zhou}(v) = \frac{d(v)}{\sum_{u \in \V}d(u)}$ and $\L_{zhou} = \I - \D_{v}^{-1/2}\H\W\D_{e}^{-1}\H^{\top}\D_{v}^{-1/2}$ where $D_{v}(u,u) = d(u) = \sum_{e \in \E}w(e)H(u,e)$ and $D_{e}(e,e) =\delta(e) = \sum_{v\in e}H(v,e)$. The forms of $\pi_{zhou}$ and $\L_{zhou}$ are exactly the same in \citet{zhou2006learning}.
\end{remark}

\begin{remark}[\citet{carletti2020random,carletti2021random}]
    When we select $\rho (\cdot)=(\cdot)^{\sigma}$ and $\Q_1=\Q_2=\H$, the probabilities of unified random walk on hypergraph are reduced to $p_1=\frac{w(e)H(u,e)(\sum_{v\in \V}H(v,e))^{\sigma+1}} {\sum_{e\in \E}w(e)H(u,e)(\sum_{v\in \V}H(v,e))^{\sigma+1}}$ and $p_2=\frac{H(v,e)}{\sum_{v\in \V}H(v,e)}$. As the special case of ours satisfying $\Q_1=\Q_2=\H$ and $\rho(\cdot)=(\cdot)^{\sigma}$, \citet{carletti2021random}'s random walk satisfies both condition~(1) and condition~(2) in Thm.~\ref{Th:EquivalencyGrpah2Hyeprgrpah}. So we obtain from Corollary~\ref{Th:stationary_distribution} that $\pi_{car}(v) = \frac{d(v)}{\sum_{u \in \V}d(u)}$ and $\L_{car} = \I - \D_{v}^{-1/2}\H\W\D_{e}^{\sigma}\H^{\top}\D_{v}^{-1/2}$ where $D_{v}(u,u) = d(u) = \sum_{e \in \E}w(e)\delta(e)^{\sigma+1}H(u,e)$ and $\D_{e}(e,e) = \delta(e) = \sum_{v\in e}H(v,e)$. The forms of $\pi_{car}$ and $\L_{car}$ are exactly the same in \citet{carletti2021random}. Note that the random walk proposed by \citet{carletti2020random} is a special case of \citet{carletti2021random} with $\sigma=1$, the $\pi$ and $\L$ are easy to obtain from our conclusions.
\end{remark}

\begin{remark}[\citet{2019random}]
When we set $\rho(\cdot)=(\cdot)^{-1} $ and $\Q_1 = \H$, the probabilities of unified random walk on hypergraph become $p_1=\frac{w(e)H(u,e)}{\sum_{e\in \E}w(e)H(u,e)}$ and $p_2=\frac{Q_2(v,e)}{\sum_{v\in \V}Q_2(v,e)}$ with an edge-dependent vertex weights $\Q_2$ in the second steps. Actually, as the special case of ours satisfying $\Q_1=\H$ and $\rho(\cdot)=(\cdot)^{-1}$.
\end{remark}

\begin{figure*}[ht]
    \centering
    \includegraphics[width=1.0\linewidth]{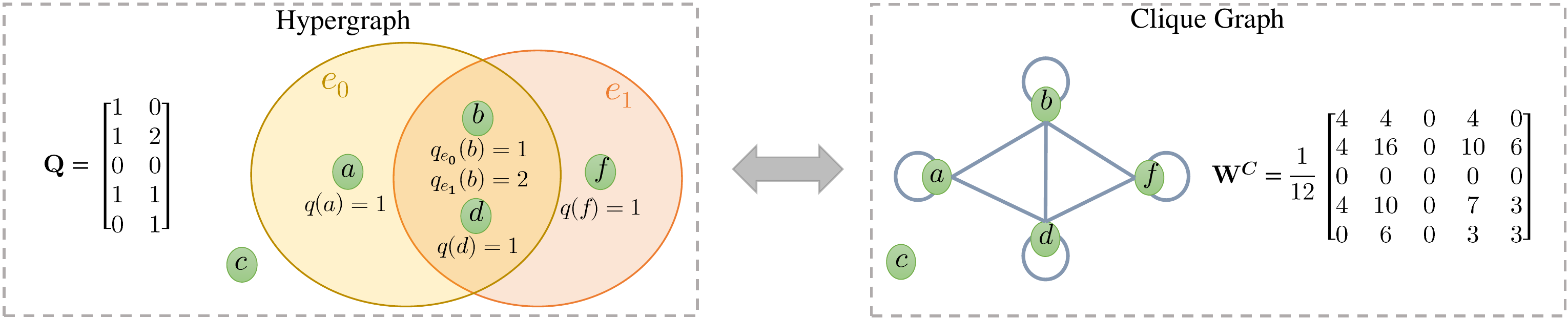}
     \caption{An example of the disconnected hypergpraph and its equivalent weighted undirected graph. $c$ is an isolated vertex. The characteristics of hypergraph are encoded in the weight incidence matrix $\Q$. The elements in $\mbf W^{C}:=\Q\D_e^{-1}\Q^{\top}$ denotes the edge-weight matrix of the clique graph.  }
    \label{fig:hypergraph2graph}
        \vspace{-2mm}
\end{figure*}

\section{Typical Hypergraph Convolution (HGNN) and Its Spectral Analysis}\label{sec:bg}
Recall Remark \ref{Remark:zhou2006}, the random walk and Laplaican proposed by \citet{zhou2006learning} are special cases~($\rho(\cdot)=(\cdot)^{-1}, \Q_1=\Q_2=\H$) of our unified random walk and unified Laplacian, respectively. Thus, now the degree of a vertex $v\in\V$ is $d(v) = \sum_{e\in \E}w(e)H(v,e)$ and for a hyperedge $e \in \E$, its degree is $\delta(e) = \sum_{v\in \V} H(v,e)$. The edge-degree matrix and vertex-degree matrix can be denoted as diagonal matrices $\D_{e} \in \R^{|\E|\xx|\E|}$ and $\D_{v} \in \R^{|\V|\xx |\V|}$ respectively. 

\subsection{Typical Spectral Hypergraph Convolution}\label{app:hgnn}
  From the seminal paper \cite{zhou2006learning}, a classical symmetric normalized hypergraph Laplacian is defined from the perspective of spectral hypergraph partition:
  $
      \L = \mathbf{I} - \D_{v}^{-1/2}\H\W\D_{e}^{-1}\H^{\top}\D_{v}^{-1/2}.
  $
  Based on it, \citet{hgnn} follow \citet{kipf2016semi} to deduce the \textit{Hypergraph Neural Network} HGNN. Specifically, they first perform an eigenvalue decomposition of $\L$: $\L = \mathbf{U}\Lambda \mathbf{U}^{\top}$, where $\Lambda$ is a diagonal matrix of eigenvalues and $\mathbf{U}$ is consisted of the corresponding regularized eigenvectors. Combining the hypergraph Laplacian with Fourier Transform of graph signal processing, they derive the primary form of hypergraph spectral convolution:
 $
    g*x = \U((\U^{\top}g)\odot (\U^{\top}x)) = \U g(\Lambda) \U^{\top}x
$
where $g(\Lambda) = diag(g(\lambda_{1}),g(\lambda_{2}),\cdots,g(\lambda_{N}))$ is a function of the eigenvalues of $\L$. To avoid expensive computation of eigen decomposition~\citep{defferrard2016convolutional}, they use  the truncated Chebyshev polynomials $T_{k}(x)$ to approximated $g(\Lambda)$, which is recursively deduced by $T_{k}(x) = 2xT_{k-1}(x) - T_{k-2}(x)$ with $T_{0}(x) = 1$ and $T_{1}(x) = x$.
$
    g*x  \approx \sum_{k=0}^{K-1} \theta_{k}T_{k}(\hat{\Lambda})x,
$
with scaled Laplacian $\hat{\Lambda} = \frac{2}{\lambda_{max}}\Lambda - \mathbf{I}$ limiting the eigenvalues in $[-1,1]$ to guarantee requirement of the Chebyshev polynomial~\citep{hammond2011wavelets}. Since the parameters in the neural network could adapt to the scaled constant~($\lambda_{max}\approx 2$), the convolution operation is further simplified to be
$
    g * x \approx \theta_{0} x-\theta_{1} \mathbf{D}_{v}^{-\frac{1}{2}} \mathbf{H} \mathbf{W} \mathbf{D}_{e}^{-1} \mathbf{H}^{\top} \mathbf{D}_{v}^{-\frac{1}{2}}x.$

However, in order to obtain the final convolutional layer similar to~\citet{kipf2016semi} proposed, \textbf{\citet{hgnn} used a specific matrix to fit the scalar parameter $\theta_0$, causing a flaw in the derivation.} In the end, they have managed to obtain a seemingly correct hypergraph spectral convolution~(HGNN):
\begin{align}
    \mbf Y = \D^{-1/2}\H\W\D_e^{-1}\H^{\top}\D_v^{-1/2}\X\mbf{\Theta},
\end{align}
where $\mbf{\Theta}$ is the parameter to be learned during the training process. There is no doubt that HGNN can work properly, but it could have low efficiency  when the hypergraph of the datasets is disconnected. Our experiments in subsection \ref{sub:citation_classification} and Figure \ref{fig:hypergraph2graph} verify our conclusion. 

\subsection{Spectral Analysis of HGNN}
We give the definition of  Rayleigh quotient to study the spectrum of Laplacian.
\begin{lemma}[ Rayleigh Quotient \citep{horn1986topics}]
    \label{le:Rayleigh_quotient}
    Assume that $\mathbf{M} \in \R^{n\xx n}$ is a real symmetric matrix and $\mbf x \in \R^{n}$ is an arbitrary nonzero vector. Let $\lambda_{1}\leq \lambda_{2}\leq \cdots \leq \lambda_{n}$ denote the eigenvalues of $\mbf M$ in order. Then the Rayleigh quotient satisfies the property:
    $$
    \lambda_{1} \leq R(\mathbf{M}, \mbf x):=\frac{\mbf x^{\top} \mathbf{M} \mbf x}{\mbf x^{\top} \mbf x} \leq \lambda_{n}.
    $$
    The equality holds that $R(\mathbf{M},\mathbf{u}_{1})=\lambda_{1}$ and $R(\mathbf{M},\mathbf{u}_{n})=\lambda_{n}$ where $\mathbf{u}_{i}$ is the eigenvector related to $\lambda_{i}, i\in \{1,n\}$. 
\end{lemma}
The Lemma \ref{le:Rayleigh_quotient} is mainly used to get all exact or approximated eigenvalues and eigenvectors of $\mathbf{M}$. 

In this part, we utilize it to prove that the smallest eigenvalue $\lambda_{min}$ of $
\L = \mathbf{I} - \D_{v}^{-\frac{1}{2}}\H\W\D_{e}^{-1}\H^{\top}\D_{v}^{-\frac{1}{2}}$ is 0 and figure out the corresponding eigenvector $\mbf u_{min}$ is $\D_{v}^{\frac{1}{2}}\mathbf{1}$.

Specifically, let $g$ denote an arbitrary column vector which assigns to each vertex $v\in \mathcal{V}$ a real value $g(v)$. We demonstrate $R(\L,g)$ as below in order to describe our conclusion distinctly:
    \begin{align*}
        \lambda_{min}\leq \frac{g^{\top} \L g}{g^{\top} g}&=\frac{g^{\top} (\mathbf{I} - \D_{v}^{-\frac{1}{2}}\H\W\D_{e}^{-1}\H^{\top}\D_{v}^{-\frac{1}{2}}) g}{g^{\top} g}\\
        &=\frac{(\D_{v}^{\frac{1}{2}}f)^{\top} (\mathbf{I} - \D_{v}^{-\frac{1}{2}}\H\W\D_{e}^{-1}\H^{\top}\D_{v}^{-\frac{1}{2}}) (\D_{v}^{\frac{1}{2}}f)}{(\D_{v}^{\frac{1}{2}}f)^{\top} (\D_{v}^{\frac{1}{2}}f)}\\
        &=\frac{f^{\top} (\D_{v}-\H\W\D_{e}^{-1}\H^{\top}) f}{f^{\top} \D_{v}f}\\
        &=\frac{\sum_{e}\sum_{v}\sum_{u}\frac{w(e)h(v,e)h(u,e)}{\delta(e)}(f(u)-f(v))^{2}}{2\sum_{v}f(v)^{2}d(v)}
    \end{align*}
    where $g=\D_v^{1/2}f$ and $f\in \R^{|\V|\xx 1}.$ It is easy to see the fact: $R(\L,g)\geq 0$, and $R(\L,g) = 0 $ if and only if $f = \mbf{1}$ (i.e. $g={\D}_{v}^{1/2}\mathbf{1}$). Then with lemma~\ref{le:Rayleigh_quotient} we get $\lambda_{min} = \min\limits_{g} R(\L,g)=R(\L,{\D}_{v}^{1/2}\mbf{1})=0$ and $\u_{min}={\D}_{v}^{\frac{1}{2}}\mbf{1}$. 
    
Actually, since the Laplacian of ~\citet{zhou2006learning} is our unified Laplacian in Corollary \ref{Th:stationary_distribution}, the property above can be directly led by Thm. \ref{th:eignvectors}.
\section{Derivation and Analysis of H-GNNs}\label{appsec:derivationModel}
\subsection{GNN induced Hypergraph NNs} \label{app:Derivation_of_GHCN}
We define $\K$ as $\Q_2\W\rho(\D_e)\Q_2^{\top}$. Then the unified hypergraph Laplacian matrix $\L$ in Thm.~\ref{Th:stationary_distribution} can be denoted as
\begin{align}
    \L = \mathbf{I}-\hat{\D}_{v}^{-1/2}\mbf K\hat{\D}_v^{-1/2},
\end{align}
\paragraph{H-GCN}
Following ~\citet{defferrard2016convolutional}, the convolutional kernel is approximated   with Chebyshev polynomial to avoid eigenvalue decomposition of $\L$.
\begin{align}
    \mbf g \star \x = \theta_0\x - \theta_1\hat{\D}^{-1/2}_{v}\mbf{K}\hat{\D}^{-1/2}_{v}\x.
\end{align}
Then we adopt the same method proposed by \cite{kipf2016semi} to set the parameter $\theta = \theta_0 = -\theta_1$ and obtain the following expression:
\begin{align}
    \mbf g \star \x = \theta\left(\I+\hat{\D}_v^{-1/2}\mbf{K}\hat{\D}_v^{-1/2}\right)\x.
\end{align}
According to Theorem \ref{th:eignvectors}, it is easy to know the eigenvalues of $\I+\hat{\D}_v^{-1/2}\mbf{K}\hat{\D}_{v}^{-1/2}$ range in $[0,2]$, which may cause the unstable numerical instabilities and gradient explosion problem. So we use the renormalization trick~\citep{kipf2016semi}:
\begin{align}
    \label{Eq:T_of_GHCN}
    \I+\hat{\D}_{v}^{-1/2}\mbf{K}\hat{\D}_{v}^{-1/2} \to \Tilde{\D}_v^{-1/2}\Tilde{\mbf{K}}\Tilde{\D}_v^{-1/2}\triangleq \Tilde{\T },
\end{align}
where $ \Tilde{\mbf K}=\mbf K+\mbf I$ can be regarded as the weighted adjacency matrix of the equivalent weighted graph with self-loops, and $\Tilde{\D}(v,v)=\sum_{u}\Tilde{\mbf K}(v,u)$ is a $|\V|\xx|\V|$ diagonal matrix. This self-loops makes the methods more robust to process disconnected hypergraphs datasets.  Finally, we drive the generalized hypergraph convolutional network(\textbf{H-GCN}):
\begin{align}
    \X^{(l+1)}=\psi(\Tilde{\mbf T}\X^{(l)}\mbf \Theta)
\end{align}

$\mbf \Theta$ is a learnable parameter and $\psi(\cdot)$ denotes an activation function.

There are several leading message-passing models of undirected graph which are easy convert to the corresponding Hypergraph version under our GHSC framework.
\paragraph{H-APPNP.} 
\begin{equation}
\begin{aligned}
\label{HAPPNP}
    &\mathbf{Z}^{(0)} = \mbf H, \mbf H = f_{\theta}(\mathbf{X});\\
    &\mathbf{Z}^{(k+1)} = (1-\alpha) \Tilde{\mbf T} \mbf Z^{(k)} +  \alpha \mbf H ;\\
    &\mbf Z^{(k)} = softmax((1-\alpha)\Tilde{\mbf T})\mbf Z^{(k-1)} + \alpha\mbf H,
\end{aligned}
\end{equation}
where $\mbf X$ is the original node feature matrix and $\mbf H$ is the feature embedding generating by a prediction neural network $f_{\theta}$. Similar with H-GCN, $\Tilde{\mbf T}$ is the symmetrical normalized weighted adjacnecy matrix of the equivalent weighted graph with self-loops.
\paragraph{H-SSGC.}
\begin{equation}
\begin{aligned}
\label{HSSGC}
\mbf Z = softmax(\frac{1}{K}\sum_{k=1}^{K}((1-\alpha)\Tilde{\mbf T}^k\mbf X + \alpha \mbf X)\mbf \Theta),
\end{aligned}
\end{equation}
where $K$ is a hyperparameter for the diffusion steps and $\mbf \Theta$ is the learnable  parameters matrix.
\paragraph{H-ChebNet.}
\begin{equation}
    \begin{aligned}
    \mbf Z = Relu(\sum_{k=0}^{K}\Tilde{\mbf T}^k \mbf X\mbf \Theta ^{(k)}),
    \end{aligned}
\end{equation}
where $\mbf \Theta ^{(k)}(k=0,\cdots,K)$ are the learnable paremeter matrix for K-th chebyshev Polynomial of $\Tilde{\mbf T}$.
\paragraph{H-GCNII.}

\begin{equation}
    \begin{aligned}
    \mbf Z^{(l+1)} = \sigma(((1-\alpha_l)\Tilde{\mbf T}\mbf Z^{(l)} + \alpha_l \mbf Z^{(0)})((1-\beta_l)\mbf I +\beta_l \mbf \Theta^{(l)})),
    \end{aligned}
\end{equation}
where $\mbf Z^{(l)}$ is the output of l-th layer of GCNII and $\mbf Z^{(0)}=f_\theta(\mbf X)78$.
\begin{remark}
Most of existing hypergraph Laplacians~\citep{zhou2006learning,carletti2020random,carletti2021random} can be viewed as special forms of the Laplacian matrix $\L$ and can be derived special  cases of  GHSC Framework.

\end{remark}

\subsection{Over-smoothing Analysis of GHSC}
\label{proof:over_smoothing}
\paragraph{Informal Definition of oversmoothing on Hypergraphs.}
In our paper, the over-smoothing issue for hypergraphs means the nodes embeddings of hypergraphs are indistinguishable. 
The equivalent clique graphs have the same vertices as hypergraphs, so the oversmoothing issue for hypergraphs is defined upon the corresponding undirected graphs.

In order to formalise the analysis of oversmoothing on Hypergraphs, we further define a quantity to measure this phenomenon as follows.
For real-world datasets, we always use non-Euclidean graphs with finite-dimensional features of vertices. It is difficult to use the absolute \textit{diffusion distance}~\citep{masuda2017random}
which measures the distance walked from starting vertex in the diffusion process. Therefore, we use the reciprocal of the mean $l_1$-norm error between $t$-step transition probability $\P^{t}$ and stationary distribution $\pi$ as the measure of over-smoothing in arbitrary starting vertex. Formally, we name the measure by \textbf{over-smoothing energy} as follows:
$$
e(i,t) = \frac{1}{\frac{1}{N}\Vert \mathbf{f}(i)\P^{t}-\pi \Vert_{1}}
$$
where $i$ denotes the sign of starting vertex $i$ and $t$ denotes the number of diffusion step. As mentioned in Corollary. \ref{Th:convergency_rate}, $\mathbf{f}(i)$ is an initial distribution which means the walker diffuse starting from vertex $i$(i.e. $\mathbf{f}(i)_{i}=1$ and $\mathbf{f}(i)_{j}=0, \forall j \neq i$ ). It is easy to observe that if the limit of $e(i,t)$ with respect to $t$ converges to infinity, it indicates the model based on $\P$ undergo severe over-smoothing issue.
Recall Corollary \ref{Th:convergency_rate}, we get a low-bounded quantity of $e(i,t)$ as $e_{low}(i,t)$:
$$
e(i,t) \geq \frac{N}{\sum_{j=1}^{N}(1-\lambda_{H})^{t}\frac{\sqrt{\hat{d}(j)}}{\sqrt{\hat{d}(i)}}} = \frac{N\sqrt{\hat{d}(i)}}{(1-\lambda_{H})^{t}
\varphi(H)}=e_{low}(i,t)
$$
where  $\varphi(H) = \sum_{j=1}^{N}\sqrt{\hat{d}(j)}$ can be seen as a constant associated with $\mathcal{H}$ which is independent with $i$ or $t$. Recall Corollary \ref{Th:convergency_rate} that $\lambda_H$ is smallest nonzero eigenvalue of $\L$.   

\paragraph{Over-smoothing analysis of H-GCN.}Here, H-GCN is used as an example of a GHSC framework derived spectral convolutional network for the analysis of oversmoothing problems. As shown in Table \ref{tab:depth_accuracy}, we can see our proposed H-GCN model also face with the over-smoothing phenomenon. Assume that we get a $K$-layers H-GCN model with the form:
$ \X^{(l+1)} = \operatorname{ReLU}(\Tilde{\mbf T}\X^{(l)}\mbf\Theta)$.
To get a simple mathematical form from above incremental convolution formula~(remove the activation layer), we could get approximated $K$-layers H-GCN as:
$$
\widetilde{\X}^{(K)} = \Tilde{\mbf T}^{K}\X^{(0)}\mbf\Theta
$$
Intuitively, from the simple form we find the \textit{over-smoothing energy} of H-GCN in vertex $i$ can be represented by $e_{low}(i,K)$. It is easy to analyze that as $K\rightarrow \infty$, $e_{low}(i,K)$ converge to infinity which implies that the \textit{over-smoothing energy} $e(i,t)$ converge to infinity. Furthermore, the convergence of \textit{over-smoothing energy} to infinity means the error between initial signal and $\pi$ converge to $0$ as the number of layers increasing, which describes the reason of over-smoothing underwent by H-GCN from a quantitative perspective. HGNN~\citep{hgnn} also suffers from over-smoothing as shown in Table \ref{tab:depth_accuracy}. Meanwhile, the theoretical analysis of over-smoothing on HGNN can be covered by the analysis of H-GCN because HGNN is a special case of H-GCN~(w/o and w re-normalization).

\section{Details of Theories and Proofs }\label{app:Missing_proofs}




\begin{definition}[Reversible Markov chain]\label{def:Reverible_markov}
    Let M be a Markov chain with state space $\mcal X$ and transition probabilities $P(u,v)$, for $u, v \in \mcal X $. We say M is
    reversible if there exists a probability distribution $\pi$ over $\mcal X$ such that
    \begin{align}
    \label{Eq:reversible_condition}
        \pi(u)P(u,v) = \pi(v)P(v,u) .   
    \end{align}
\end{definition}
\begin{lemma}[\citet{2019random}]
    \label{Le:markov2undigraph}
    Let $M$ be an irreducible Markov chain with finite state space $\mcal S$ and transition probabilities $P(u,v)$ for $u,v \in \mcal S$. $M$ is reversible if and only if there exists a weighted, undirected graph $\mcal G$ with vertex set $\mcal S$ such that a random walk on $\mcal G$ and $M$ are equivalent.
\end{lemma}
\paragraph{Proof.} Let $\pi$ be the stationary distribution of $M$~(suppose that is irreducible). Note that $\pi(u)\neq 0$ due to the irreducibility of $M$. Let $\G$ be a graph with vertices $\mcal S$. Different from \citet{2019random}, we set the edge weights of $\G$ to be
\begin{align}
    \label{Eq:weights_of_equivelent_undigraph}
    \omega(u,v)=c\pi(u)P(u,v),\quad \forall\ u,v\in \mcal S
\end{align}
where $c>0$ is a  constant for normalizing $\pi$. With reversibility, $\G$ is well-defined~(i.e. $\omega(u,v)=\omega(v,u)$). In a random walk on $\G$, the transition probability from $u$ to  $v$ in one time-step is: 
$\frac{\omega(u,v)}{\sum_{v\in \V}\omega(u,v)}=\frac{c \pi(u)P(u,v)}{\sum_{v\in \V} c\pi(u)P(u,v)}=P(u,v)$, since $\sum_{w\in \mcal S}P(u,w)=1$. 
Thus, if $M$ is reversible, recall Definition \ref{def:equi}, the stated claim holds. The other direction follows from the fact that a random walk on an undirected graph is always reversible (Aldous \& Fill,[70]).

\subsection{Proof of Lemma \ref{le:EquivalencyUndigrpah2Hyeprgrpah}}\label{proof:le:EquivalencyUndigrpah2Hyeprgrpah}
\begin{restatable}{lemma}{restatLemmaOne}
\label{le:EquivalencyUndigrpah2Hyeprgrpah}
    Let $\mathcal{H}(\V,\E,\W,\Q_1,\Q_2)$ denote the generalized hypergraph in Definition \ref{def:generalized hypergraph}.
    Let $\mathcal{F}_{(Q_1,Q_2)}(u,v):=\sum_{e \in \E}w(e)\rho(\delta(e))Q_1(u,e)Q_2(v,e)$ and $\mathcal{T}_{(Q)}(u):=\sum_{e\in\E}w(e)\delta(e)\rho(\delta(e))Q(u,e)$. When $\Q_1,\Q_2$ satisfies the following equation
    \vspace{-1mm}
    \begin{snugshade}
        \vspace{-5mm}
    \begin{align}
    \label{Eq:equivalencyEquation}
       & \mathcal{T}_{(Q_2)}(u) \mathcal{T}_{(Q_1)}(v) \mathcal{F}_{(Q_1,Q_2)}(u,v) \notag
       \\ &= 
        \mathcal{T}_{(Q_2)}(v) \mathcal{T}_{(Q_1)}(u) \mathcal{F}_{(Q_1,Q_2)}(v,u), \forall u,v\in \V
    \end{align}
    \end{snugshade}
    \vspace{-2mm}
   there exists a weighted undirected clique graph $\G^{C}$ such that a random walk on $\mathcal{H}$ is equivalent to a random walk on $\G^{C}$ with edge weights $\omega(u,v) = {\mathcal{T}_{(Q_2)}(u)}\mathcal{F}_{(Q_1,Q_2)}(u,v)/{\mathcal{T}_{(Q_1)}(u)}$ ~if~ $\mathcal{T}_{(Q_1)}(u)\neq 0$ and 0 otherwise.
   And the symmetrical Laplacian of $\G^{C}$ is 
    \begin{align}
        L(u,v) = \mathcal{I}_{\{u=v\}}-\mathcal{T}_{(Q_2)}(u)^{-{1}/{2}}\omega(u,v)\mathcal{T}_{(Q_2)}(v)^{-{1}/{2}}
    \end{align}
    where $\mathcal{I}_{\{\cdot\}}$ denotes the indicator function.
\end{restatable}
This Lemma suggests a direction to explore the equivalency condition. Any generalized hypergraph composed of the vertex weights $\Q_1,\Q_2$ satisfying \Eqref{Eq:equivalencyEquation} is equivalent to the undirected graph.

\paragraph{Proof.}
Form Lemma~\ref{Le:markov2undigraph}, we just need to prove Markov chain with state space $\V$ is reversible.
The transition probabilities of the unified random walk on $\H$ are:
\begin{align*}
    P(u,v) &= \sum_{e \in \E}\frac{w(e)\rho(\delta(e))Q_1(u,e)Q_2(v,e)}{d(u)}
    = \frac{\mathcal{F}_{(Q_1,Q_2)}(u,v)}{\mathcal{T}_{(Q_1)}(u)}
\end{align*}
By \Eqref{Eq:equivalencyEquation}, we have:
\begin{align*}
      \mathcal{T}_{(Q_2)}(u) \frac{\mathcal{F}_{(Q_1,Q_2)}(u,v)}{\mathcal{T}_{(Q_1)}(u)} =\mathcal{T}_{(Q_2)}(v) \frac{\mathcal{F}_{(Q_1,Q_2)}(v,u)}{\mathcal{T}_{(Q_1)}(v)} , 
\end{align*}
i.e. 
\begin{align*}
      \mathcal{T}_{(Q_2)}(u) P(u,v) =\mathcal{T}_{(Q_2)}(v) P(v,u) , 
\end{align*}
Comparing with ~\Eqref{Eq:reversible_condition}, we normalize $G(Q_2)$ to define the stationary distribution of $M$:
\begin{align}
\label{eq:quivalentUndigraphStationaryDistribution}
    \pi(u) := \frac{\mathcal{T}_{(Q_2)}(u)}{\sum_{u\in \V}\mathcal{T}_{(Q_2)}(u)}, \forall u\in \V,
\end{align}
which can be easy to verify by $\sum_{u}\pi(u)P(u,v)=\pi(v)$. Thus, $\pi(u)P(u,v)=\pi(v)P(v,u)$, which suggests that $M$ is reversible. By Lemma~\ref{Le:markov2undigraph}, the  edge weight $\omega(u,v)$ of $\G^{C}$ is 
\begin{align*}
    \omega(u,v) = c\pi(u)P(u,v) =  \mathcal{T}_{(Q_2)}(u) P(u,v) = \frac{\mathcal{T}_{(Q_2)}(u)}{\mathcal{T}_{(Q_1)}(u)}\mathcal{F}_{(Q_1,Q_2)}(u,v)
\end{align*}
To gain the Laplacian of $\G^{C}$, we calculate the sum of row of edge weights matrix,
\begin{align*}
    \sum_{v\in \V} \omega(u,v) = \frac{\mathcal{T}_{(Q_2)}(u)}{\mathcal{T}_{(Q_1)}(u)} \sum_{v\in \V} \mathcal{F}_{(Q_1,Q_2)}(u,v) = \mathcal{T}_{(Q_2)}(u)
\end{align*}
Then we get the symmetrical Laplacian of $\G^C$:
\begin{align}
\label{eq:equivalentUndigraphLaplacian}
    L(u,v) = \mathcal{I}_{\{u=v\}}- \mathcal{T}_{(Q_2)}(u)^{-\frac{1}{2}}\omega(u,v)\mathcal{T}_{(Q_2)}(v)^{-\frac{1}{2}}    
\end{align}
\subsection{Proof of Theorem \ref{Th:EquivalencyGrpah2Hyeprgrpah}}\label{proof:Th:EquivalencyGrpah2Hyeprgrpah}
\restathmtwo*
\paragraph{Proof.}
\paragraph{1) Proof based on Lemma \ref{le:EquivalencyUndigrpah2Hyeprgrpah}  }
For condition (1), 
\begin{align}
\label{eq:edgeIndependentF}
\mathcal{F}_{(Q_1,Q_2)}(u,v) &=\sum_{e \in \E}w(e)\rho(\delta(e))Q_1(u,e)Q_2(v,e)=q_1(u)q_2(v)\sum_{e \in \E}w(e)\rho(\delta(e))H(u,e)H(v,e)  \notag\\
\mathcal{T}_{(Q_i)}(u)&=\sum_{e\in\E}w(e)\delta(e)\rho(\delta(e))Q_i(u,e)= q_i(u)\sum_{e\in\E}w(e)\delta(e)\rho(\delta(e))H(u,e), ~i=\{1,2\}    
\end{align}
Substituting the above equations into \Eqref{Eq:equivalencyEquation}, then \Eqref{Eq:equivalencyEquation} holds.

For condition (2),
\begin{align}
    \label{eq:edgeDependentF}
    \mathcal{F}_{(Q_1,Q_2)}(u,v)&=\sum_{e \in \E}w(e)\rho(\delta(e))Q_1(u,e)Q_2(v,e)=k\sum_{e \in \E}w(e)\rho(\delta(e))Q_2(u,e)Q_2(v,e) \notag \\
    \mathcal{T}_{(Q_1)}(u)&=\sum_{e\in\E}w(e)\delta(e)\rho(\delta(e))Q_1(u,e)= k\sum_{e\in\E}w(e)\delta(e)\rho(\delta(e))Q_2(u,e)= k\mathcal{T}_{(Q_2)}(u)
\end{align}
Similarly, substituting the above equations into \Eqref{Eq:equivalencyEquation}, then \Eqref{Eq:equivalencyEquation} holds.

\paragraph{2
) Proof based on Lemma \ref{Le:markov2undigraph} }

I)~ For Condition~(1):\\
Because $\Q_1$ and $\Q_2$ are both edge-independent, 
we set $Q_1(u,e)=q_1(u)$ and $Q_{2}(v,e) = q_2(v)$ for all hyperedge $e$.
From Lemma \ref{Le:markov2undigraph}, the key is to prove the unified hypergraph random walk on $\mcal H$ under condition~(1) is reversible~(see Definition \ref{def:Reverible_markov}). It is hard to find the explicit form of $\pi$ before we get Corollary \ref{cor:indepen-sysme}. Fortunately, by Kolmogorov's criterion, that is equal to prove:
$$ p_{v_{1}, v_{2}} p_{v_{2}, v_{3}} \cdots p_{v_{n}, v_{1}}=p_{v_{1}, v_{n}} p_{v_{n}, v_{n-1}} \cdots p_{v_{2}, v_{1}}$$
for arbitrary subset $\{v_{1},\cdots,v_{n}\} \subseteq \V$.
The transition probabilities of the generalized random walk on $\H$ are:
\begin{align*}
    p(u,v) &= \sum_{e \in \E}\frac{w(e)\rho(\delta(e))Q_1(u,e)Q_2(v,e)}{d(u)}
    =  \frac{q_1(u)q_2(v)}{d(u)}\sum_{e \in \E}w(e)\rho(\delta(e))H(u,e)H(v,e)\\
    & = \frac{q_1(u)q_2(v)}{d(u)}\sum_{e \in E(u,v)}w(e)\rho(\delta(e))
\end{align*}
where $E(u,v)=\{e\in\E:u\in e,~v\in e\}$ to be the set of hyperedges incident to both $v$ and $u$. Then we have:
\begin{align*}
     &p_{v_{1}, v_{2}}\cdots p_{v_{n}, v_{1}} \\
     &= \left(\frac{q_1(v_{1})q_2(v_{2})}{d(v_{1})} \sum_{e \in E\left(v_{1}, v_{2}\right)} w(e)\rho(\delta(e))\right)\cdots \left(\frac{q_1(v_{n})q_2(v_{1})}{d(v_{n})} \sum_{e \in E\left(v_{n}, v_{1}\right)} w(e)\rho(\delta(e))\right)\\
     &=\left(\prod_{i=1}^{n}\frac{q_1(v_{i})q_2(v_{i})}{d(v_{i})}\right)\cdot\left(\prod_{i=1}^{n}\sum_{e \in E\left(v_{i}, v_{i+1}\right)} w(e)\rho(\delta(e))\right),~\text{where~set}~v_{n+1} = v_{1}\\
     &= \left(\frac{q_1(v_{1})q_2(v_{n})}{d(v_{1})} \sum_{e \in E\left(v_{n}, v_{1}\right)} w(e)\rho(\delta(e))\right)\cdots\left(\frac{q_1(v_{2})q_2(v_{1})}{d(v_{2})} \sum_{e \in E\left(v_{1}, v_{2}\right)} w(e)\rho(\delta(e))\right)\\
     &=p_{v_{1}, v_{n}} \cdots p_{v_{2}, v_{1}}
\end{align*}
So the unified random walk on $\mcal H$ under condition (1) is reversible.

As a matter of fact, we can succinctly obtain the reversibility via the stationary distribution. Specifically, we can verify the stationary distribution is 
\begin{align}
    \pi(u) =\frac{\sum_{e\in \E}w(e)\delta(e)\rho(\delta(e))Q_2(u,e)}{\sum_{v\in\V}\sum_{e\in \E}w(e)\delta(e)\rho(\delta(e))Q_2(v,e)}
\end{align}
by $\sum_{u}\pi(u)P(u,v)=\pi(v)$. Then with $\pi(u)P(u,v)=\pi(v)P(v,u)$ holding, we know our unified random walk on  $\mcal H$ under condition (1) is reversible. Since $\mcal H$ is connected, the unified random on $\mcal H$ is irreducible.
From Lemma.\ref{Le:markov2undigraph}, We know that the current unified random walk on $\mcal H$ is equivalent to a random walk on an undirected graph $\G$ with vertex set $\V$. By \Eqref{Eq:weights_of_equivelent_undigraph}, the edge weights of $\G$ are: 
\begin{align}
    \label{Eq:condition1weights}
    \omega(u,v) = c\pi(u)P(u,v)=\sum_{e\in \E}w(e)\rho(\delta(e))Q_2(u,e)Q_2(v,e)    
\end{align}
where $c=\sum_{v\in\V}\sum_{e\in \E}w(e)\delta(e)\rho(\delta(e))Q_2(v,e)$. Note that $P(u,v)>0$ if and only if $\omega(u,v)>0$, so $\G$ is the clique graph of $\mcal H$.

II) For condition (2) that $\Q_1=k\Q_2$,\\
we have $\P = k\D_v^{-1}\Q_2 \W \rho(\D_{e})\Q_2^{\top}$. Thanks to the underlying symmetry adjacency matrix of $\P$~(i.e. $\Q_2 \W \rho(\D_{e})\Q_2$ is symmetrical), we have: $\mbf{1}^{\top}\hat{D}_{v}\P=\mbf{1}^{\top}\hat{D}_{v}^{\top}$. Thus, the stationary distribution is

\begin{align}
    \pi(u) = \frac{\hat{d}(u)}{\sum_{v}\hat{d}(v)} = \frac{\sum_{e\in \E}w(e)\delta(e)\rho(\delta(e))Q(u,e)}{\sum_{v\in\V}\sum_{e\in \E}w(e)\delta(e)\rho(\delta(e))Q(v,e)}
\end{align}
 Further, we would prove that the unified random walk on $\mcal H$ under current condition~(2) is reversible.
\begin{align*}
    \pi(u) p({u,v}) &= \frac{\hat{d}(u)}{\sum_{v'}\hat{d}(v')}\sum_{e\in \E} \frac{w(e)\rho(\delta(e))Q(u,e)Q(v,e)}{d(u)}\\
    &= \frac{d(v)}{\sum_{v'}d(v')}\sum_{e\in \E} \frac{w(e)\rho(\delta(e))Q(v,e)Q(u,e)}{d(v)} = \pi(v) p({v,u})
\end{align*}
So the unified hypergraph random walk on $\mcal H$ under condition (2) is reversible. Since $\mcal H$ is connected, the unified random on $\mcal H$ is irreducible. From Lemma.\ref{Le:markov2undigraph}, We know that the current unified random walk is equivalent to a random walk on a weighted, undirected graph $\G$ with vertex set $\V$. By \Eqref{Eq:weights_of_equivelent_undigraph}, the edge weights of $\G$ are:
\begin{align}
    \label{Eq:condition2weights}
    \omega(u,v) = c\pi(u)P(u,v)=\sum_{e\in \E}w(e)\rho(\delta(e))Q(u,e)Q(v,e)    
\end{align}
where $c=\sum_{v'}d(v')=\sum_{v\in\V}\sum_{e\in \E}w(e)\delta(e)\rho(\delta(e))Q(v,e)$. Note that $P(u,v)>0$ if and only if $\omega(u,v)>0$, so $\G$ is the clique graph of $\mcal H$.
On the whole, we get the specific equivalent weighted undigraphs under our condition (1) and condition (2). Note that they have the same edge weights expression~( \Eqref{Eq:condition1weights},\Eqref{Eq:condition2weights}):
$$
\W^{C} = \Q_{2}\W\rho(\D_{e})\Q_{2}^{\top}
$$
which further suggests the connection between condition (1) and condition~(2) stated in Corollary \ref{cor:indepen-sysme} and Thm. \ref{Th:stationary_distribution}. 

\paragraph{Remark:} Let $\rho(\cdot)=(\cdot)^{-1}$ and $\W=\mbf{I}$, we can get the undigraph shown in Figure \ref{fig:hypergraph2graph_connected}.

\begin{figure*}[h]
    \centering
    \includegraphics[width=1.0\linewidth]{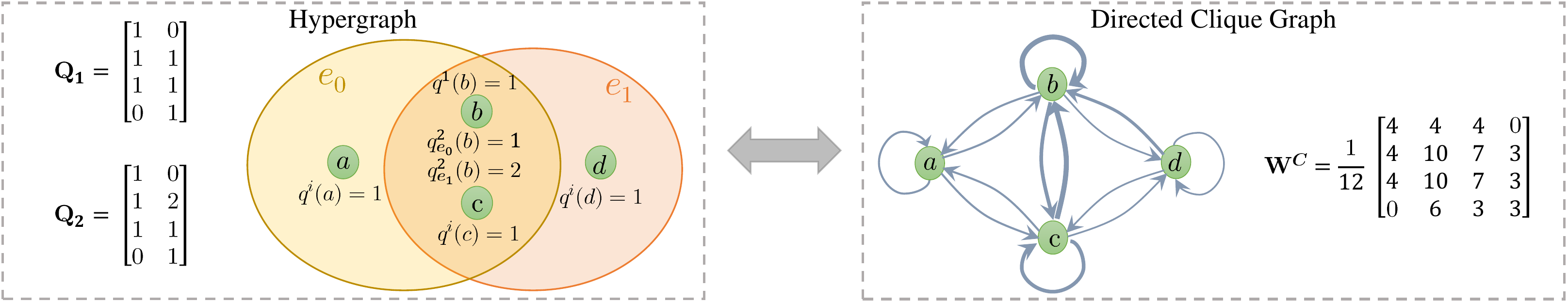}
     \caption{An example of hypergraph and its equivalent weighted directed graph, where $q^i(\cdot)=Q_i(\cdot,e), i\in\{1,2\}$. Here, $\Q_1$ is edge-independent and $\Q_2$ is edge-dependent. The characteristics of the hypergraph are encoded in the weighted incidence matrices~(i.e. vertex weights) $\Q_1, \Q_2$.
    $\W^{C}:=\Q_1\D_e^{-1}\Q_2^{\top}$ denotes the edge-weight matrix of the clique graph  
     and can be viewed as the embedding of high-order relationships.}
    \label{fig:hypergraph_irreversible}
        \vspace{-2mm}
\end{figure*}
\subsection{Nonequivalent Condition. }\label{app:Nonequivalent_condition}
\begin{theorem}
\label{Th:UnquivalencyGrpah2Hyeprgrpah}
    There at least exists one generalized hypergraph $\mathcal{H}(\V,\E,\W,\Q_1,\Q_2)$ (Definition \ref{def:generalized hypergraph}) with edge-dependent weights and $\Q_1\neq k\Q_2$, such that a random walk on $\mathcal{H}$ is not equivalent to a random walk on its undirected clique graph $\G^{C}$ for any choice of edge weights on $\G^{C}$.
\end{theorem}
\paragraph{Remark.}

Theorem \ref{Th:UnquivalencyGrpah2Hyeprgrpah} tells that the random walk on a hypergraph can not be always equivalent to a random walk on an undigraph, and open up the possibility to construct a hypergraph is not equivalent to an undigraph, which may inspire researcher to further study for higher-order interactions and other insurmountable bottlenecks on hypergraphs.
\paragraph{Proof.} 
When the vertex weights are edge-dependent, the reversibility is not always satisfied in our unified random walk on a hypergraph.
By Lemma~\ref{Le:markov2undigraph}, if the unified random walk is not time-reversible, it could not equivalent to random walks on digraphs for any choice of edge weights. 

A irreversible example can see Figure \ref{fig:hypergraph_irreversible}.
The probability transition matrix of the unified random walk on the left hypergraph~(with $\W=\I, \rho(\cdot)=(\cdot)^{-1}$) is given below:
\begin{align}
\label{Eq:transition}
    \P=\D_v^{-1}\Q_1 \W \rho(\D_{e})\Q_2^{\top} = \D_v^{-1}\Q_1 \D_{e}^{-1}\Q_2^{\top}=\begin{bmatrix}
 \frac{1}{3}  &\frac{1}{3}  &\frac{1}{3}  &0 \\[1mm]
 \frac{1}{6}& \frac{5}{12} & \frac{7}{24}  &\frac{1}{8}  \\[1mm]
  \frac{1}{6}& \frac{5}{12} & \frac{7}{24}  &\frac{1}{8}  \\[1mm]
  0& \frac{1}{2}  &\frac{1}{3}   &\frac{1}{3} 
\end{bmatrix},
\end{align}
 and the stationary distribution of the unified random walk on hypergraph with $\P$ is:
 \begin{align}
    \pi =  [\frac{3}{17},\frac{7}{17},\frac{5}{17},\frac{2}{17}]
 \end{align}
We verify $\pi(1)P'(1,2)=\frac{3}{17}\xx \frac{1}{3}\neq\frac{7}{17}\xx \frac{1}{6}=\pi(2)P(2,1)$, which indicates that the Markov chain represented by the unified random walk in Figure \ref{fig:hypergraph_irreversible} with transition probabilities $\P$ 
is irreversible.
 
\subsection{Failure Condition of Function.}
\label{proof:pro:k-uniform}
As a matter of fact, the $\rho$ in our random walk framework can not always work and we give its failure condition as follows.
\begin{proposition}
    	Let $\mathcal{H}(\V,\E,\W,\mathbf{Q})$ associated with the unified random walk in Definition \ref{def:random_walk_Definition}.
    	$\rho(\cdot)$ fails to work in our unified random walk if the hyperedge degrees are edge-independent~(i.e. $\delta(e)=\delta(e'),\forall e' \in \E$).
\end{proposition}
\paragraph{Proof.}
Given $\delta(e)$ are edge-independent, let us suppose $\delta(e)=k$, then
\begin{align*}
     P(u,v)&=\sum_{e \in \E} \frac{w(e)Q_1(u,e)Q_2(v,e)\rho(\delta(e))} {d(u)}
     =\sum_{e \in \E} \frac{w(e)Q_1(u,e)Q_2(v,e)\rho(\delta(e))} {\sum_{e \in \E} w(e)\delta(e)\rho(\delta(e))Q_1(v,e)}\\
     &=\sum_{e \in \E} \frac{w(e)Q_1(u,e)Q_2(v,e)\rho(k)} {\sum_{e \in \E}
     w(e)\delta(e)\rho(k)Q_1(v,e)} 
     =\sum_{e \in \E} \frac{\rho(k) w(e)Q_1(u,e)Q_2(v,e)} {\rho(k) \sum_{e \in \E} w(e)\delta(e)Q_1(v,e)} \\
      &=\sum_{e \in \E} \frac{w(e)Q_1(u,e)Q_2(v,e)} {\sum_{e \in \E} w(e)\delta(e)Q_1(v,e)} 
\end{align*}
which indicates $P(u,v)$ is independent to $\rho(\cdot)$, i,e. $\rho(\cdot) $ fails to work.

\par
It is obvious that the $k$-uniform hypergraph has edge-independent hyperedge degrees. Formally, we have the following proposition.
\begin{proposition}
    \label{pro:k-uniform}
	 Let $\mathcal{H}(\V,\E,\W,\Q_1,\Q_2)$ denote the generalized hypergraph in Definition \ref{def:generalized hypergraph} with $\rho (\cdot)=(\cdot)^{\sigma}$ and $\Q_1=\Q_2=\H$ , then $\mathcal{H}$ is a $k$-uniform hypergraph(i.e each hyperedge contains the same number of vertices) if and only if $ P(u,v)$ in~\Eqref{Eq:tansition_propobility} is independent with $\sigma$.
		\end{proposition}
\paragraph{Proof.}

	For $\Q = \mathbf{H}$, then $\delta(e)=\sum_{v \in e}h(v,e)=k$.\\
	So, for all $u,v\in \V$, we have:
	\begin{align}
		\label{la}
		K^{(\sigma)}(u,v):=
		\sum_{e \in \E}w(e)(\delta(e))^{\sigma}q_{e}(u)q_{e}(v)
		 &=k^{\sigma}\sum_{e \in \E}w(e)H(u,e)H(v,e)
	\end{align}
	Then we get the entries of the transition matrix $\P^{(\sigma)}$:
	\begin{align}
	P^{(\sigma)}(u,v)&:= \frac{K^{(\sigma)}(u,v)}{\sum_{b}K^{(\sigma)}(u,b)}
	=\frac{k^{\sigma}\sum_{e \in \E}w(e)H(u,e)H(v,e)}{\sum_{b\in \V}k^{\sigma}\sum_{e \in \E}w(e)H(u,e)H(b,e)}\\
	&=\frac{\sum_{e \in \E}w(e)H(u,e)H(v,e)}{\sum_{b\in \V}\sum_{e \in \E}w(e)H(u,e)H(b,e)}
\end{align}
It is easy to see that $ P^{(\sigma)}(u,v)$ is independent with $\sigma$.\\
The other direction, we find when $ P^{(\sigma)}(u,v)$ in is independent with $\sigma$,then:

\begin{align}
	&\frac{\partial P^{(\sigma)}(u,v) }{\partial \sigma}=\frac{\frac{\partial d(u)}{\partial \sigma}K^{(\sigma)}(u,v)-\frac{\partial K^{(\sigma)}(u,v)}{\partial \sigma}d(u) }{d^{2}(u)}\\
	&= \frac{\sum_{b}\sum_{e}\sum_{e'}w(e)w(e')h(u,e)h(u,e')h(v,e')h(b,e)\delta(e)^{\sigma}\delta(e')^{\sigma}(\ln(\delta(e))-\ln(\delta(e')))}{d^{2}(u)}\\
	&=0, \forall ~u,v \in \V
\end{align}
It is equal to $\exists k \in \mathbf{Z}^{++}$ s.t.:
$$\delta(e)=\delta(e')=k~for~all~e\in \E$$.i.e. the hypergraph is k-uniform.
With the condition that $\Q_1=\Q_2=\H$, our unified random walk on hypergraph reduces to a random walk on hypergraph that leaving only incident relationships between hyperedges and vertices. Actually, the reduced random walk mentioned is a common structure in real-world applications. The proposition is to say when the hypergraph based on the reduced random walk is uniform, the function $\rho(\cdot)$ does not impact the importance between vertices any more. 

\subsection{Proof of Corollary  \ref{Th:stationary_distribution}.}\label{proof:Th:stationary_distribution}

Before proving the Corollary \ref{Th:stationary_distribution}, we further illustrate the fact that Lemma \ref{cor:indepen-sysme} states.

\begin{restatable}{lemma}{restacorotowpointone}
    \label{cor:indepen-sysme}
   Let $\mathcal{H}(\V,\E,\W,\Q_1,\Q_2)$ be the generalized hypergraph in Definition \ref{def:generalized hypergraph}. 
    Given a $\mcal H$ satisfying condition (1) in Thm. \ref{Th:EquivalencyGrpah2Hyeprgrpah}, there exists a $\mcal H'$ satisfying the condition (2) such that the random walk on $\mcal H$ is equivalent to that  on $\mcal H'$.

\end{restatable}
\paragraph{Proof of Lemma~\ref{cor:indepen-sysme}.}
When the condition (2) in Thm. \ref{Th:EquivalencyGrpah2Hyeprgrpah} holds~(i.e. $\Q_1=\Q_2$), the transition probabilities of the unified random walk on $\mcal H$ are
\begin{small}
\begin{align}
\label{Eq:transition_of_condition2}
    P^{c2}(u,v) = \frac{\sum_{e \in \E}w(e)\rho(\delta(e))Q_{2}(u,e)Q_{2}(v,e)}{\sum_{e\in\E}w(e)\rho(\delta(e)\delta(e)Q_{2}(u,e)}
\end{align}
\end{small}

Meanwhile, if the condition (1) in Thm. \ref{Th:EquivalencyGrpah2Hyeprgrpah} holds~(i.e. $\Q_1,\Q_2$ are both edge-independent), for all hyperedge $e$, we represent $Q_{1}(u,e)$, $Q_{2}(v,e)$ as $q_{1}(u)$, $q_2(v)$ respectively, and the transition probabilities are 
\begin{small}
\begin{align}
    \label{Eq:transition_of_condition1}
    P^{c1}(u,v) &= \frac{\sum_{e \in \notag \E}w(e)\rho(\delta(e))Q_{1}(u,e)Q_{2}(v,e)}{\sum_{e\in\E}w(e)\rho(\delta(e)\delta(e)Q_{1}(u,e)} \notag
    = \frac{q_1(u)q_2(v)\sum_{e \in \E}w(e)\rho(\delta(e))H(u,e)H(v,e)}{q_1(u)\sum_{e\in\E}w(e)\rho(\delta(e)\delta(e)H(u,e)}\\ 
    &=\frac{q_2(u)q_2(v)\sum_{e \in \E}w(e)\rho(\delta(e))H(u,e)H(v,e)}{q_2(u)\sum_{e\in\E}w(e)\rho(\delta(e)\delta(e)H(u,e)}
    =\frac{\sum_{e \in \E}w(e)\rho(\delta(e))Q_{2}(u,e)Q_{2}(v,e)}{\sum_{e\in\E}w(e)\rho(\delta(e)\delta(e)Q_{2}(u,e)}
\end{align}
\end{small}
 It can be observed that $P^{c1}(u,v)= P^{c2}(u,v), \forall u,v\in \V$. By definition \ref{def:equi}, the Lemma \ref{cor:indepen-sysme} holds.

Next, based on  Lemma \ref{cor:indepen-sysme}, we prove the following main conclusion.
\restathmthree*

\paragraph{Proof of Corollary~\ref{Th:stationary_distribution}.} 
When the generalized hypergraph satisfies Theorem \ref{Th:EquivalencyGrpah2Hyeprgrpah}, we obtain the \Eqref{eq:edgeIndependentF} and \Eqref{eq:edgeDependentF} in Appendix ~\ref{proof:le:EquivalencyUndigrpah2Hyeprgrpah}. substituting the equations into \Eqref{eq:equivalentUndigraphLaplacian} and \Eqref{eq:quivalentUndigraphStationaryDistribution}, we can obtain the Corollary~\ref{Th:stationary_distribution}.

\subsection{Spectral Range of Laplacian Matrix}\label{proof:th:eignvectors}
 The following Theorem proves the upper bound for eigenvalues of $\L$ which meets the requirement of Chebyshev polynomials to enable the derivation for our proposed H-GCN(a simple case of GHSC) in Appendix \ref{app:Derivation_of_GHCN} and leads to Corollary .~\ref{Th:convergency_rate} in the paper.

\begin{theorem}
 \label{th:eignvectors}
  Let $\L = \mathbf{I}-\hat{\D}_{v}^{-1/2}\Q_2\mathbf{W}\rho(\D_e)\Q_2^{\top}\hat{\D}_{v}^{-1/2}$ denote the unified hypergraph Laplacian matrix in Corollary \ref{Th:stationary_distribution}
and $\lambda_{1}\leq \lambda_{2}\leq \cdots \leq \lambda_{n}$ denote the eigenvalues of $\L$ in order.
\begin{itemize}
    \item [1)]$\lambda_{min} = \lambda_{1} = 0$ and $\mbf{u}_{1} = \hat{\D}_{v}^{\frac{1}{2}}\mathbf{1}$(the eigenvector associated with $\lambda_{1}$ )
    \item [2)]For $k = 2,3,\cdots,n$, we have
    $$
        \lambda_{k} = \inf\limits_{f \perp \hat{\D}_{v}^{\frac{1}{2}} S_{k-1}} \frac{\sum_{e,u,v}\beta(e,u,v)(f(u)-f(v))^{2}}{\sum_{v} f(v)^{2} d(v)},
    $$
    here, \ $S_{k-1}$ is the subspace spanned by eigenvectors $\{\u_1,\cdots,\u_{k-1}\}$ where $\u_{i}$ related to $\lambda_{i}$ and $\beta(e,u,v) = w(e)\rho(\delta(e))Q_2(v,e)Q_2(u,e)$.
    \item[3)]$\lambda_{max}=\lambda_{n} \leq 2$
\end{itemize}
\end{theorem}
\paragraph{Proof:}
To analyze the generalized Laplacian proposed in our work, we use the Rayleigh quotient mentioned in the Lemma \ref{le:Rayleigh_quotient} to give out some conclusions about the eigenvalues and eigenvectors of the Laplacian.\\
1) Deduce the Rayleigh quotient of $\L_{2}$:
\begin{align*}
     \frac{g^{\top} \L g}{g^{\top} g}&=\frac{g^{\top} (\mathbf{I} - \hat{\D}_{v}^{-\frac{1}{2}}\Q_2\W\rho(\D_{e})\Q_2^{\top}\hat{\D}_{v}^{-\frac{1}{2}}) g}{g^{\top} g}\\
        &=\frac{(\hat{\D}_{v}^{\frac{1}{2}}f)^{\top} (\mathbf{I} - \hat{\D}_{v}^{-\frac{1}{2}}\Q_2\W\rho(\D_{e})\Q_2^{\top}\hat{\D}_{v}^{-\frac{1}{2}}) (\hat{\D}_{v}^{\frac{1}{2}}f)}{(\hat{\D}_{v}^{\frac{1}{2}}f)^{\top} (\hat{\D}_{v}^{\frac{1}{2}}f)}\\
        &=\frac{f^{\top} (\hat{\D}_{v}-\Q\W\rho(\D_{e})\Q^{\top}) f}{f^{\top} \hat{\D}_{v}f}\\
        &=\frac{\sum_{u}\sum_{v}\sum_{e}w(e)\rho(\delta(e))Q_2(v,e)Q_2(u,e)(f(u)-f(v))^{2}}{2\sum_{v}f(v)^{2}\hat{d}(v)}
\end{align*}
where $g=\hat{\D}_{v}^{1/2}f$ and $f\in \R^{|\V|\xx 1}$. It is easy to see the fact: $R(\L,g)\geq 0$ and $R(\L,g) = 0 $ if and only if $f = \mbf{1}$ (i.e. $g=\hat{\D}_{v}^{1/2}\mathbf{1}$), Then with Lemma~\ref{le:Rayleigh_quotient} we get $\lambda_{1}=\lambda_{min} = \min\limits_{g} R(\L,g)=R(\L,\hat{\D}_{v}^{1/2}\mbf{1})=0$ and $\u_{1}=\hat{\D}_{v}^{\frac{1}{2}}\mbf{1}$.\\
2) From Courant–Fischer theorem~\citep{horn1986topics}, we have:
\begin{align*}
\lambda_{k} &= \inf\limits_{g \perp S_{k-1}} R(\L, g)=\inf\limits_{g \perp S_{k-1}}  \frac{g^{\top} \L g}{g^{\top} g}\\
&=\inf\limits_{f \perp \hat{\D}_{v}^{\frac{1}{2}}S_{k-1}}\frac{\sum_{u}\sum_{v}\sum_{e}w(e)\rho(\delta(e))Q_2(v,e)Q_2(u,e)(f(u)-f(v))^{2}}{2\sum_{v}f(v)^{2}\hat{d}(v)}
\end{align*}
3) We have the fact:
$$(f(u)-f(v))^{2}\leq 2(f^{2}(u)+f^{2}(v))$$
And from Lemma \ref{le:Rayleigh_quotient}, we get:
\begin{align*}
    \lambda_{n}&=\lambda_{max}=R(\L,\u_{max}) \notag \\
    & = \inf\limits_{f \perp \hat{\D}_{v}^{\frac{1}{2}}S_{n-1}}\frac{\sum_{u}\sum_{v}\sum_{e}w(e)\rho(\delta(e))Q_2(u,e)Q_2(v,e)(f(u)-f(v))^{2}}{2\sum_{v}f(v)^{2}\hat{d}(v)} \notag\\
    &\leq\inf\limits_{f \perp \hat{\D}_{v}^{\frac{1}{2}}S_{n-1}} \frac{\sum_{u}\sum_{e}w(e)\rho(\delta(e))\delta(e)Q_2(u,e)f(u)^2+\sum_{v}\sum_{e}w(e)\rho(\delta(e))\delta(e)Q_2(v,e)f(v)^2}{2\sum_{v}f(v)^{2}\hat{d}(v)} \notag \\ 
    &\leq \inf\limits_{f \perp \hat{\D}_{v}^{\frac{1}{2}}S_{n-1}} \frac{2\sum_{u}\hat{d}(u)f^{2}(u)}{\sum_{v}\hat{d}(v)f^{2}(v)}=2\\
\end{align*}
\subsection{Proof of Corollary \ref{Th:convergency_rate}} \label{proof:Th:convergency_rate}
\begin{restatable}[]{corollary}{restathmfour}
    \label{Th:convergency_rate}
    Let $\mathcal{H}(\V,\E,\W,\Q_1,\Q_2)$ be the generalized hypergraph in Definition \ref{def:random_walk_Definition}. When $\mcal H$ satisfies any of two conditions in Thm. \ref{Th:EquivalencyGrpah2Hyeprgrpah}, let $\L$ and $\pi$ be the hypergraph Laplacian matrix and stationary distribution from Corollary \ref{Th:stationary_distribution}.
    Let $\lambda_{H}$ denote the smallest nonzero eigenvalue of $\L$. 
    Assume an initial distribution $\mbf{f}$ with $f{(i)} = 1$ ($f{(j)}=0,\forall ~j \neq i$) which means the corresponding walk starts from vertex $v_{i}$. Let $\mbf p^{(k)} = \mbf{f}\P^{k}$ be the probability distribution after $k$ steps unified random walk where $\P$ denotes the transition matrix, then $p^{(k)}{(j)}$ denotes the probability of finding the walker in vertex $v_{j}$ after $k$ steps. We have: 
    \begin{align}
    \small
        \left|p^{(k)}(j)-\pi(j)\right| \leq \sqrt{\frac{\hat{d}(j)}{\hat{d}(i)}}(1-\lambda_{H})^{k}.
    \end{align}
\end{restatable}
\paragraph{Proof:}
From ~\Eqref{Eq:transition_matrix}:
$$\P = \hat{\D}_{v}^{-1}\Q \W \rho(\D_{e})\Q^{\top} = \hat{\D}_{v}^{-\frac{1}{2}}(\I-\L)\hat{\D}_{v}^{\frac{1}{2}}$$
Then we have:
\begin{align*}
      \mbf{1}^{\top}\hat{\D}_{v}\P=\mbf{1}^{\top}\hat{\D}_{v}
\end{align*}
Then the stationary distribution $\pi\in\R^{1\xx|\V|}$ of $\P$ can be denoted as:
\begin{align}
 \mbf{\pi} = \frac{\mbf{1}^{\top}\hat{\D}_{v}}{c}   
\end{align}
, where $c=\mbf{1}^{\top}\hat{\D}_{v}\mbf{1}=\sum_{v\in\V}\hat{d}(v)> 0$ is the sum of $\mbf{1}^{\top}\hat{\D}_{v}$.
Let $\lambda_{1}\leq \lambda_{2}\leq \cdots \leq \lambda_{n}$ denote the eigenvalues of $\mbf L$ in order. Then we assume that $\mbf{f}\hat{\D}_{v}^{-\frac{1}{2}} = \sum\limits_{i=1}^{n}a_{i}\Tilde{\mbf{u}}^{\top}_{i}\in\R^{1\xx|\V|}$, where $\Tilde{\mbf{u}}_i\in\R^{|\V|\xx 1}$ denotes the $l_{2}$-norm~orthonormal eigenvector related with $\lambda_{i}$. From Thm .\ref{th:eignvectors}, we know $\Tilde{\mbf{u}}_{1}=\frac{\mbf{u}_1}{\|\mbf u_1\|_2}=\frac{\hat{\D}_{v}^{\frac{1}{2}}\mathbf{1}}{\sqrt{c}}$. Then $a_{1}$ can be computed as:
$$a_{1} =\frac{\Tilde{\mbf{u}}_{1}^{\top}(\mbf{f}\hat{\D}_{v}^{-\frac{1}{2}})^{\top}}{\Tilde{\mbf{u}}_{1}^{\top}\cdot \Tilde{\mbf{u}}_{1}}=\Tilde{\mbf{u}}_{1}^{\top}(\mbf{f}\hat{\D}_{v}^{-\frac{1}{2})})^{\top}=\frac{(\hat{\D}_{v}^{\frac{1}{2}}\mathbf{1})^{\top} (\mbf{f}\hat{\D}_{v}^{-\frac{1}{2}})^{\top}}
{\sqrt{c}}=\frac{1}{\sqrt{c}}$$ 
with the fact that $\mbf{f}\in\R^{1\xx|\V|}$ is a probability distribution(i.e. $\mbf{f\mbf{1}}=1$).\\
Then,
\begin{align*}
\left|p_{j}(k)-\pi_{j}\right| &= \left|(\mbf{f}\P^{k})_{j}-\pi_{j} \right|
= \left|(\mbf{f}\P^{k}-\frac{\mbf{1}^{\top}\hat{\D}_{v}}{c})_j \right| 
=\left|(\mbf{f}\P^{k}-a_{1}\Tilde{\mbf{u}}_{1}^{\top}\hat{\D}_{v}^{\frac{1}{2}})_{j} \right|  \\
&=\left|(\mbf{f}\hat{\D}_{v}^{-\frac{1}{2}}(\I-\L)^{k}\hat{\D}_{v}^{\frac{1}{2}}-a_{1}\Tilde{\mbf{u}}_{1}^{\top}\hat{\D}_{v}^{\frac{1}{2}})_{j} \right| 
=\left|(\sum\limits_{i=1}^{n}a_{i}\Tilde{\mbf{u}}^{\top}_{i}(\I-\L)^{k}\hat{\D}_{v}^{\frac{1}{2}}-a_{1}\Tilde{\mbf{u}}_{1}^{\top}\hat{\D}_{v}^{\frac{1}{2}})_{j} \right|\\
&=\left|(\sum\limits_{i=1}^{n}a_{i}\Tilde{\mbf{u}}^{\top}_{i}(1-\lambda_{i})^{k}\hat{\D}_{v}^{\frac{1}{2}}-a_{1}\Tilde{\mbf{u}}_{1}^{\top}\hat{\D}_{v}^{\frac{1}{2}})_{j} \right|
=\left|(\sum\limits_{i=2}^{n}a_{i}\Tilde{\mbf{u}}^{\top}_{i}(1-\lambda_{i})^{k}\hat{\D}_{v}^{\frac{1}{2}})_{j} \right| \\
&\leq (1-\lambda_H)^k\sqrt{\hat d(j)}\left|(\sum_{i=1}^na_i\Tilde{\mbf{u}}_i^{\top})_j\right| 
= (1-\lambda_H)^k\sqrt{\hat d(j)}\left|(\mbf{f\hat{D}}_v^{-1/2})_j \right| \\
&\leq (1-\lambda_{H})^{k}\frac{\sqrt{\hat{d}(j)}}{\sqrt{\hat{d}(i)}}
\end{align*}

\section{Details of Experiments}\label{sec:Details_of_Experiments}
Note that our H-GNNs are based on the Laplacian led by the equivalent condition in Thm .\ref{Th:EquivalencyGrpah2Hyeprgrpah}, which means the vertex weights we use in our experimental datasets should satisfy the condition (1) or condition (2) in Thm. \ref{Th:EquivalencyGrpah2Hyeprgrpah}. 

\subsection{Hyper-parameter Strategy}
\label{sec:hyper-parameter}
We use grid search strategies to adjust the hyper-parameters of our H-GCN and H-SSGC. The range of hyper-parameters listed in Table \ref{tab:hyper-parameter-citation_object}, Table \ref{tab:hyper-parameter-fold}, Table \ref{tab:hyper-parameter-QA12}, and Table \ref{tab:hyper-parameter-QA13}.

\begin{table}[htbp] 
\centering
\caption{Hyper-parameter search range for citation network classification and visual object classification.}
\vspace{1mm}
\resizebox{0.5 \linewidth}{!}{
\begin{tabular}{c|c|c}
\toprule
     Methods & Hyper-parameter        & Range               \\ 
\midrule
    \multirow{10}{*}{\begin{tabular}[c]{@{}l@{}}H-GNNs\end{tabular}}
    & $\sigma$ & \{-2,-1,-0.5,0,0.5,1,2\} \\
    & $\gamma$(visual object classification)    & \{0.1, 0.2, 0.4, 0.5, 0.8,1.0\} \\
    & Learning rate        & \{0.001, 0.005, 0.01\}                  \\
    & Hidden dimension & \{128\} \\
    & Layers & \{2,4,6,8,16,32,64\} \\
    & Weight decay & \{1e-3,1e-4,5e-4, 1e-5\}  \\
    & Dropout rate        & \{0.1,0.2,0.3, 0.4, 0.5\}  \\            
    & Optimizer & Adam  \\
    & Epoch & 1000 \\
    & Early stopping patience & 100 \\
    & GPU & Tesla V100 \\
    
\bottomrule                     
\end{tabular}                           
}
\label{tab:hyper-parameter-citation_object}
\end{table}

\begin{table}[ht] 
\centering
\caption{Hyper-parameter search range for H-GCN and H-GCNII in fold classification.}
\vspace{1mm}
\resizebox{0.58 \linewidth}{!}{
\begin{tabular}{l |l|l l}
\toprule
    Methods & Hyper-parameter        & Range         & best      \\ 
\midrule
    \multirow{14}{*}{\begin{tabular}[c]{@{}l@{}}H-GCN\end{tabular}}
    & Amino-acids type input size     & \{32,64,128\}  &  64 \\
    & Secondary-structure input size & \{32,64,128\}  &  64 \\
    & Readout layer input size & \{256,512,1024\} & 1024 \\
    & Layers L & \{2,3,4,5,6\} & 4 \\
    & Batch size & \{64,128,256\} & 128 \\
    & Weight decay & \{1e-3,1e-4, 1e-5\} & 1e-3\\
    & Learning rate        & \{0.0001,0.001, 0.005, 0.01\} &    0.0001              \\
    & Dropout rate        & \{0.1,0.2,0.3, 0.4, 0.5\}     &  0.2           \\
    & $\sigma$ & \{-1\}  & -1\\
    & $\gamma$ & \{0.1,0.2,,0.4,0.5,0.8,1.0\} & 0.1 \\
    & Optimizer & Adam & - \\
    & Epochs & 300 & - \\
    & Early stopping patience & 30 & -\\
    & GPU & GTX  2080Ti & - \\
    \midrule
    \multirow{16}{*}{\begin{tabular}[c]{@{}l@{}}H-GCNII\end{tabular}}    
    & Amino-acids type input size     & \{64,128,256\}  & 256 \\
    & Secondary-structure input size & \{64,128,256\}  &  256 \\
    & Readout layer input size & \{256,512,1024\} & 256 \\
    & Layers L & \{2,4,6,8,12,16,32\} & 2 \\
    & Batch size & \{64,128,256\} & 256 \\
    & Weight decay & \{1e-3,1e-4, 1e-5\} & 1e-5\\
    & Learning rate        & \{0.0001,0.001, 0.005, 0.01\} &    0.001              \\
    & Dropout rate        & \{0.1,0.2,0.3, 0.4, 0.5\}     &  0.3           \\
    & $\sigma$ & \{-1\}  & -1\\
    & $\gamma$ & \{0.1,0.2,,0.4,0.5,0.8,1.0\} & 0.1 \\
    & $\alpha$ & \{0.1\} & 0.1 \\
    & $\beta$ & \{ 0.5 \} & 0.5 \\
    & Optimizer & Adam & - \\
    & Epochs & 300  & -\\
    & Early stopping patience & 30  & -\\
    & GPU & Tesla V100  & -\\
\bottomrule                     
\end{tabular}                           
}
\label{tab:hyper-parameter-fold}
\end{table}

\begin{table}[ht] 
\centering
\caption{Hyper-parameter search range for H-GCN and H-SSGC in protein Quality Assessment~(CASP10,CASP11,CASP13 for training, CASP12 for testing).} 
\vspace{1mm}
\resizebox{0.8 \linewidth}{!}{
\begin{tabular}{l| l|l l}
\toprule
     Methods & Hyper-parameter        & Range           & best    \\ 
\midrule
    \multirow{14}{*}{\begin{tabular}[c]{@{}l@{}}H-GCN\end{tabular}}
    & Amino-acids type input size     & \{128,256,512\}  &  512 \\
    & Secondary-structure input size & \{32,64,128\}  &  64 \\
    & Readout layer input size & \{256,512,1024\} & 1024 \\
    & Layers L & \{2,3,4,5,6\} & 4 \\
    & Batch size & \{64,128,256\} & 64 \\
    & Weight decay & \{1e-3,1e-4, 1e-5\} & 1e-5\\
    & Learning rate        & \{0.0001,0.001, 0.005, 0.01\} &    0.0001              \\
    & Dropout rate        & \{0.1,0.2,0.3, 0.4, 0.5\}     &  0.5           \\
    & $\sigma$ & \{-1\}  & -1\\
    & $\gamma$ & \{0.1,0.2,,0.4,0.5,0.8,1.0\} & 0.1 \\
    & Optimizer & Adam   & - \\
    & Epochs & 200  & - \\
    & Early stopping patience & 20  & - \\
    & GPU & Tesla P40  & - \\
    \midrule
    \multirow{16}{*}{\begin{tabular}[c]{@{}l@{}}H-SSGC\end{tabular}}    
    & Amino-acids type input size     & \{128,256,512\}  & 512 \\
    & Secondary-structure input size & \{32,64,128\}  &  128 \\
    & Readout layer input size & \{256,512,1024\} & 256 \\
    & Layers L & \{2,4,6,8,12,16,18,32\} & 18 \\
    & Batch size & \{64,128,256\} & 128 \\
    & Weight decay & \{1e-3,1e-4, 1e-5\} & 1e-5\\
    & Learning rate        & \{0.0001,0.001, 0.005, 0.01\} &    0.0005              \\
    & Dropout rate        & \{0.1,0.2,0.3, 0.4, 0.5\}     &  0.3           \\
    & $\sigma$ & \{-1\}  & -1\\
    & $\gamma$ & \{0.1,0.2,,0.4,0.5,0.8,1.0\} & 0.2 \\
    & $\alpha$ & \{1,0.97,0.95,0.9,0.85,0.8,0.75,0.7,0.65,0.6\} & 0.65 \\
    & $\beta$ & \{ 1,0.95,0.90,0.85,0.8 \} & 0.95 \\
    & Optimizer & Adam   & -\\
    & Epochs & 200  & -\\
    & Early stopping patience & 20 & - \\
    & GPU & Tesla P40  & -\\
    
\bottomrule                     
\end{tabular}  
}
\label{tab:hyper-parameter-QA12}
\end{table}

\begin{table}[ht] 
\centering
\caption{Hyper-parameter search range for H-GCN model in protein Quality Assessment~(CASP10-12 for training, and CASP13 for testing).} 
\vspace{1mm}
\resizebox{0.7 \linewidth}{!}{
\begin{tabular}{l| l|l l}
\toprule
     Methods & Hyper-parameter        & Range           & best    \\ 
\midrule
    \multirow{14}{*}{\begin{tabular}[c]{@{}l@{}}H-GCN\end{tabular}}
    & Amino-acids type input size     & \{128,256,512\}  &  128 \\
    & Secondary-structure input size & \{32,64,128\}  &  64 \\
    & Readout layer input size & \{256,512,1024\} & 1024 \\
    & Layers L & \{2,3,4,5,6\} & 4 \\
    & Batch size & \{64,128,256\} & 1024 \\
    & Weight decay & \{1e-3,1e-4, 1e-5\} & 1e-5\\
    & Learning rate        & \{0.0001,0.001, 0.005, 0.01\} &    0.001              \\
    & Dropout rate        & \{0.1,0.2,0.3, 0.4, 0.5,0.6,0.7\}     &  0.7           \\
    & $\sigma$ & \{-1\}  & -1\\
    & $\gamma$ & \{0.1,0.2,,0.4,0.5,0.8,1.0\} & 0.1 \\
    & Optimizer & Adam   & - \\
    & Epochs & 200 & - \\
    & Early stopping patience & 20 & - \\
    & GPU & Tesla P40 & - \\
    \midrule
    \multirow{16}{*}{\begin{tabular}[c]{@{}l@{}}H-SSGC\end{tabular}}    
    & Amino-acids type input size     & \{128,256,512\}  & 128 \\
    & Secondary-structure input size & \{32,64,128\}  &  64 \\
    & Readout layer input size & \{256,512,1024\} & 1024 \\
    & Layers L & \{2,4,6,8,12,16,18,32\} & 8 \\
    & Batch size & \{64,128,256\} & 1024 \\
    & Weight decay & \{1e-3,1e-4, 1e-5\} & 1e-5\\
    & Learning rate        & \{0.0001,0.001, 0.005, 0.01\} &    0.001              \\
    & Dropout rate        & \{0.1,0.2,0.3, 0.4, 0.5\}     &  0.2           \\
    & $\sigma$ & \{-1\}  & -1\\
    & $\gamma$ & \{0.1,0.2,,0.4,0.5,0.8,1.0\} & 0.2 \\
    & $\alpha$ & \{1,0.97,0.95,0.9,0.85,0.8,0.75,0.7,0.65,0.6\} & 0.65 \\
    & $\beta$ & \{ 1,0.95,0.90,0.85,0.8 \} & 0.95 \\
    & Optimizer & Adam   & -\\
    & Epochs & 200  & -\\
    & Early stopping patience & 20  & -\\
    & GPU & Tesla P40  & -\\
    
\bottomrule                     
\end{tabular}  
}
\label{tab:hyper-parameter-QA13}
\end{table}
Extra parameter settings for Visual Object Classification include $k=6$.

\subsection{Baselines of Spectral Convolutions}\label{sec:baselines}
For HGNN~\citep{hgnn}, in addition to the flaw during the derivation (i.e. use a specific matrix to fit the scalar parameter $\theta_0$), when doing the Visual Object Classification experiment in their public code, it also takes Gaussian kernel as the incidence matrix for hypergraph learning, which lacks theoretical guarantee. So in our experiment, we just use the incidence matrix $\H$ which only consists of elements $\{0,1\}$ as the incidence matrix using in HGNN. 

HGAT and  HNHN we use the same grid search strategy as our methods.

\subsection{Citation Network Classification}
\paragraph{Datasets.}
The datasets we use for citation network classification include co-authorship and co-citation datasets: PubMed, Citeseer, Cora~\citep{sen2008collective} and DBLP~\citep{rossi2015network}. We adopt the hypergraph version of those datasets directly from  \citet{hypergcn}, where
hypergraphs are created on these datasets by assigning each document as a node and each hyperedge
represents (a) all documents co-authored by an author in co-authorship dataset and (b) all documents
cited together by a document in co-citation dataset.  The initial features of each document (vertex) are represented by bag-of-words features. The details about vertices, hyperedges and features are shown in Table~\ref{tab:dataset_hypergcn}.

\begin{table}[htbp]
\centering
\caption{Real-world hypergraph datasets used in our citation network classification task.}
\renewcommand\tabcolsep{5.0pt} 
\label{tab:dataset_hypergcn}
\vspace{-0mm}
\scalebox{0.95}{
\begin{tabular}{l|c|c|c|c|c}
\toprule
Dataset& \# vertices & \# Hyperedges & \# Features & \# Classes &  \# isolated vertices\\
\midrule
\textbf{Cora}~(co-authorship)
& 2708         & 1072         &1433   & 7     & 320(11.8\%)   \\

\textbf{DBLP}~(co-authorship)
& 43413        & 22535        & 1425       & 6     & 0 (0.0\%)     \\

\textbf{Pubmed}~(co-citation)
& 19717        & 7963         & 500        & 3      & 15877 (80.5\%)   \\

\textbf{Cora}~(co-citation)
& 2708         & 1579         & 1433       & 7     & 1274 (47.0\%)     \\

\textbf{Citeseer}~(co-citation)
& 3312         & 1079         & 3703       & 6     & 1854 (55.9\%)     \\
\bottomrule
\end{tabular}
}
\end{table}

\paragraph{Settings and baselines.}

 We adopt the same dataset and train-test splits (10 splits) as provided by in their publically available implementation\footnote{https://github.com/malllabiisc/HyperGCN, Apache License}. Note that this dataset just has the edge-independent vertex weights $\H$, which is also called the incidence matrix. So this experiment can be regarded as a special case of the specific application of our model~(i.e. $\Q=\H$).  
 
 For baselines MLP+HLR, HNHN~\citep{dong2020hnhn}~, HyperSAGE~\cite{arya2020hypersage}, UniGNN~\citep{UniGNN}, HGNN~\citep{hgnn} and FastHyperGCN~\citep{hypergcn}, HyperGCN~\citep{hypergcn}, UniGNN~\citep{UniGNN} we reuse the results reported by \citet{UniGNN}.
 For HNHN~\cite{dong2020hnhn} and HGAT~\cite{HyperGAT}, we implement them according to their public code.
 
We use cross-entropy loss and Adam SGD optimizer with early stopping with patience of 100 epochs to train GHSC. The key hyper-parameter like $\alpha$ $\beta$ in H-GCNII or other H-GNNs, we follow the settings in those GNN papers.
 For other hyper-parameters, we use the grid search strategy.
 More details of hyper-parameters can be found in Table \ref{tab:hyper-parameter-citation_object}.

 \subsection{Visual Object Classification} \label{app:sub:object_classification}
\begin{table}[htbp]
\centering
\caption{summary of the ModelNet40 and NTU datasets}

\vspace{-3mm}
\scalebox{0.8}{
    \begin{tabular}{c|c|c}
    \toprule
    \text { Dataset } & \text { ModelNet40 } & \text { NTU } \\
    \midrule
    \text { Objects } & 12311 & 2012 \\
    \text { MVCNN Feature } & 4096 & 4096 \\
    \text { GVCNN Feature } & 2048 & 2048 \\
    \text { Training node } & 9843 & 1639 \\
    \text { Testing node } & 2468 & 373 \\
    \text { Classes } & 40 & 67 \\
    \bottomrule
    \end{tabular}
}
\vspace{-2mm}
\label{tab:object_dataset}
\end{table}

\paragraph{Datasets and Settings.}
We employ two public benchmarks:  Princeton ModelNet40
dataset~\citep{wu20153d} and the National Taiwan University
(NTU) 3D model dataset~\citep{chen2003visual}, as shown in Table \ref{tab:object_dataset}.

In this experiment, each 3D object is represented by the feature vectors which are extracted by Multi-view Convolutional Neural Network (MVCNN) \citep{su2015multi} or Group-View Convolutional Neural Network (GVCNN) \citep{feng2018gvcnn}. The features generated by different methods can be considered as multi-modality features. The hypergraph structure we designed is similar to \citet{zhang2018dynamic}(
but they did not give spectral guarantees for supporting the rationality of their practices). We represent the hypergraph structure as a edge-dependent vertex weight $\Q$ to satisfy the condition (2) in Thm. \ref{Th:EquivalencyGrpah2Hyeprgrpah}~(i.e. $\Q_1=\Q_2=\Q$). Specifically, we firstly generate hyperedges by $k$\textit{-NN} approach, i.e. each time one object can be selected as centroid and its $k$ nearest neighbors are used to generate one hyperedge including the centroid itself~(in our experiment, we set $k=10$). Then, given the features of data, the  vertex-weight matrix $\mathbf{Q}$ is defined as 
\begin{align}
    \Q(v,e)=\left\{\begin{array}{cl}
	\operatorname{exp}(\frac{-d(v,v_c)}{\gamma\hat{d}^2}), &\text{if } v\in e  \\
	0, & \text{otherwise},
\end{array}\right.
\end{align}

where $d(v,v_c)$ is the euclidean distance of features between an object $v$ and the centroid object $v_c$ in the hyperedge and $\hat{d}$ is the average distance between objects. $\gamma$ is a hyper-parameter to control the flatness.
Because we have two-modality features generated by MVCNN and GVCNN, we can obtain the matrix $\Q_{\{i\}}$ which corresponds to the data of the $i$-th modality~($i\in\{1,2\}$). After all the hypergraphs from different features have been generated, these matrices $\Q_{\{i\}}$ can be concatenated to build the multi-modality hypergraph matrix $\Q=[\Q_{\{1\}},\Q_{\{2\}}]$. The features generated by GVCNN or MVCNN can be singly used, or concatenated to a multi-modal feature for constructing the hypergraphs. 

\par For baselines, we just compare with HGNN method for multi-modality learning, following the settings of Appendix \ref{sec:baselines}. We also compare our methods using two-modality features to recent SOTA methods on ModelNet40 dataset.  And we use the datasets provided by its public Code~\footnote{https://github.com/iMoonLab/HGNN, MIT License}. 
We use cross-entropy loss and Adam SGD optimizer with early stopping with patience of 100 epochs to train H-GNNs. 
More details of hyper-parameters can be found in Table \ref{tab:hyper-parameter-citation_object}. 

\paragraph{\textcolor{blue}{Additional Results on ModelNet40 Dataset.}}
Table \ref{tab:ModelNet40_accuracy} depicts that our methods significantly outperform the image-input or point-input methods. These results demonstrate that our methods can capture the similarity of objects in the feature space to improve the performance of the classification task. 
\begin{table}[thbp]

\def\p{$\pm$} 
\centering
\setlength\tabcolsep{12 pt} 
\vspace{-2mm}
\caption{ Classification accuracy  (\%) on  ModelNet40. The \textit{embedding} means the output representations of MVCNN+GVCNN Extractor.}
\vspace{-4mm}
\scalebox{0.85}{
\begin{tabular}{l|cc}
    \toprule 
    \multirow{1}{*} { Methods } & \multirow{1}{*} { input }  & Accuracy \\
    \midrule
    MVCNN \citep{feng2018gvcnn} &  image  &  90.1  \\
    PointNet \citep{qi2017pointnet} &  point &  89.2  \\
    PointNet++ \citep{qi2017pointnet++} &  point &  90.1  \\
    DGCNN~\citep{wang2019dynamic} &  point &  92.2  \\
    InterpCNN~\citep{mao2019interpolated} & point & 93.0 \\
    SimpleView~\citep{uy2019revisiting} & image & 93.6 \\
    pAConv~\citep{xu2021paconv} & point & 93.9 \\
    \hline
    HGAT~\citep{HyperGAT} & embedding & 96.4 \\
    UniGNN~\citep{UniGNN} & embedding & 96.7 \\
    HGNN~\cite{hgnn} &  embedding &  97.2 \\
    \hline
    H-ChebNet &  embedding &  97.0 \\
    H-SSGC &  embedding &  97.1 \\
    H-APPNP & embedding & 97.2   \\
    H-GCN &    embedding & 97.3   \\
    H-GCNII &    embedding  &  \textbf{ 97.8}  \\
    \bottomrule
\end{tabular}
}
    \vspace{-3.5mm}
    \label{tab:ModelNet40_accuracy}    
\end{table}

\subsection{Protein Quality Assessment and Fold Classification}\label{subsec:QA_Fold}

\paragraph{Motivation for constructing Protein Hypergraphs.} Protein fold classification is vital for mining the property of proteins and studying the relationship between protein structure and function, and protein evolution.  The function of a protein is primarily determined by its 3D structure, which contains the high-order interaction of amino-acids. However, most researchers in protein learning represent a protein as a sequence or a simple graph ~\citep{baldassarre2020graphqa,hermosillaintrinsic} that does not model the higher-order interactions between amino acids well. 
Inspired by~\citet{maruyama2001learning}, who construct the protein hypergraph to solve the conformation problem, we represent the protein as an EDVW-hypergraph to explore the high-order relationship between amino-acids for obtaining a comprehensive protein representation via H-GNNs.
 
\paragraph{Protein hypergraph modeling.}

At the high level, a protein is a chain of amino acids (residues) that will form 3D structure by spatial folding.
In order to simultaneously consider protein sequence and spatial structure information, we build sequence hyperedge and distance hyperedge. Specifically, given a protein with $|S|$ amino acids, we choose $\tau$ consecutive amino acids $(v_i,v_{i+1},\cdots,v_{i+\tau})$ to connect to form a sequence hyperedge and choose amino acids whose spatial Euclidean distance is less than a threshold $\epsilon>0$ to connect to form a spatial hyperedge, where $v_i(i=1,\cdots,|S|)$ represent the $i$-th amino acid in the sequence. Let $\E_s$ and $\E_e$ denote sequence hyperedge and spatial hyperedge, respectively.
Then, we design an edge-dependent vertex-weight matrix $\Q$ for capturing the more granular high-order relationships of proteins  below~(actually, our models H-GCN and H-SSGC allow one to design a more comprehensive $\Q$ for learning proteins better):
\begin{align}
    Q(v,e)=\left\{\begin{array}{cl}
	1, &\text{if } v\in e\operatorname{ and } e\in \E_s  \\
	\operatorname{exp}(\frac{-d(v,v_c)}{\gamma\hat{d}_{v_c}^2}), &\text{if } v\in e \operatorname{ and } e\in \E_e \\
	0, & \text{otherwise}
\end{array}\right.
\end{align}
where $d(v,v_c)<\epsilon$ is the euclidean distance between an amino acid $v$ and the centroid amino acid $v_c$ in the hyperedge and $\hat{d}_{v_c}$ is the average distance between $v_c$ and the other amino acids $\{v_i\}_{i\neq c}$. $\gamma$ is a hyper-parameter to control the flatness. Here we just design an edge-dependent vertex-weights matrix $\Q$ for satisfying the condition (2) in Thm. \ref{Th:EquivalencyGrpah2Hyeprgrpah}~(i.e. $\Q_1=\Q_2=\Q$). Note that this $\Q$ matrix pays more attention to the information of the sequence. 

\paragraph{Experimental settings}
In our experiment, we set $\tau=6$ and $\epsilon=8\mathring{A}$. The initial node features, following \citet{baldassarre2020graphqa}, are composed of amino acid types and 3D spatial features including dihedral angles, surface accessibility and secondary structure type generated by DSSP~\citep{kabsch1983dictionary}.
Note that both Quality Assessment and Fold Classification include graph-level tasks, which means we should add global pooling layers to readout the node representations from H-GNNs. 

Here, for the sake of simplicity, we adopt the permutation-invariant operator \textit{mean pooling} and a \textit{single layer MLP} as \textbf{Readout} layers to obtain a global hypergraph embedding and add the \textit{softmax} (classification) or \textit{sigmoid}~(regression) activation function before output.

\subsubsection{Protein Fold Classification}
\paragraph{Datasets and settings}
 The dataset that we used for training, validation and test is SCOPe 1.75 data set of \citet{hou2018deepsf}. This dataset includes 16,712 proteins covering 7 structural classes with total of 1195 folds. The 3D structures of proteins are obtained from SCOpe 1.75 database \citep{murzin1995scop}, in which each protein save in a PDB file. The datasets have three test sets: 1) Fold, where proteins from the same superfamily do not appear in the training set; 2) Family, in which proteins from the same family are not present in the training set; 3) Family, where proteins from the same family that are present in the training set.
 
 For baselines, we include sequence-based methods pre-trained unsupervised on millions of protein sequences: \citet{rao2019evaluating,bepler2018learning,strodthoff2020udsmprot}, 3D structure based model: \citet{kipf2016semi, diehl2019edge, baldassarre2020graphqa} and \citet{gligorijevic2020structure}, who process the sequence with LSTM first and then apply GCNN. The accuracy of the above baselines is reused from \citet{hermosillaintrinsic} reported. We adopt Adam optimizer to minimize the cross-entropy loss of our methods. Hyper-parameters search range can see Table \ref{tab:hyper-parameter-fold}.
 
\begin{table*}[thbp]
\def\p{$\pm$} 
\setlength\tabcolsep{6pt} 
\centering
\vspace{-0mm}
\caption{Comparison of our method to others on protein Quality Assessment tasks. At the residue level, We report \textit{Pearson correlation} across all residues of all decoys of all targets~($R$) and \textit{Pearson correlation} all residues of per decoys and then average all decoys~($R_{decoy}$) with LDDT scores. At the global level, we report \textit{Pearson correlation} across all decoys of all targets~($R$) and \textit{Pearson correlation} per target and then average over all targets~($R_{target}$) with GDT$\_$TS scores. }
    \vspace{-2mm}
\scalebox{0.88}{
\begin{tabular}{c|c|cc|cc}
    \toprule 
    \multirow{2}{*} { Test set } & \multirow{2}{*} { Methods } &
      \multicolumn{2}{c|} { GDT$\_$TS } & \multicolumn{2}{c} { LDDT } \\
    & & $R$  &  $R_{target}$ & $R$ & $R_{model}$ \\
     
    \hline
    \multirow{3}{*}{\begin{tabular}[c]{@{}l@{}}CASP13\end{tabular}} 
    & HGNN &  0.714 &  0.622  &  0.651  &  0.337  \\
    & H-GCN~(ours) & \textbf{ 0.718 } &  0.620 &  \textbf{ 0.657}  &  0.352  \\
     & H-SSGC~(ours) &  0.628  &  0.565  &  0.654  & \textbf{ 0.435}   \\
    \hline
    \multirow{8}{*}{\begin{tabular}[c]{@{}l@{}}CASP12\end{tabular}} 
    & VoroMQA &  - &  0.557  &  -  &  -  \\
    & RWplus &  - &  0.313  &  -  &  -   \\
    & 3D CNN        & -   &  0.607  &  -  &  -    \\
    & AngularQA &  0.651 &  0.439  &  -  &  -   \\
    \cline{2-6}
    & HGNN & 0.667  &\textbf{ 0.582}  & 0.632    &     0.319  \\
    & H-GCN (ours) &  0.737  &  \textbf{0.609} &   {0.656}  &  0.340     \\
    & H-SSGC (ours) &  \textbf{0.760}  &  0.554  &  \textbf{0.678}  &  \textbf{0.449}   \\
    \bottomrule
\end{tabular}
}
    \vspace{-2mm}
    \label{tab:Protein_QA_fold_12_13}    
\end{table*}

\subsubsection{\textcolor{blue}{Protein Quality Assessment (QA).}}

We add this additional experiments for providing a new tasks to evaluate the hypergraph algorithms and verfy our framework can be used for Protein QA tasks.

Protein QA is used to estimate the quality of computational protein models in terms of divergence from their native structure. It is a regression task aiming to predict how close the decoys to the unknown, native structure.

Inspired by \citet{baldassarre2020graphqa}, we train our models on Global Distance Test Score~\citep{zemla2003lga}, which is the global-level score, and the Local Distance Difference Test \citep{mariani2013lddt}, an amino-acids-level score. The loss function of QA is defined as the Mean Squared Error~(MSE) losses:
\begin{align}
\label{Eq:QAloss}
     \mathcal{L}_g = MSE(\mathcal{P}_{pred}^{g} - \operatorname{GDT\_TS})\qquad
     \mathcal{L}_l = \sum_{i=1}^{|S|}MSE(\mathcal{P}_{pred_i}^{l} - \operatorname{LDDT}_i)
\end{align}
where $\mathcal{P}^{g}_{pred}$ and $\mathcal{P}^{l}_{pred}$ denote predicted score of global and local respectively.

\begin{table}[htbp]
\centering
\vspace{-2mm}
\caption{summary of CASP datasets}
\vspace{-0mm}
\scalebox{0.88}{
    \begin{tabular}{c|cc|c}
    \toprule
    \text { Dataset } & \text { Targets } & \text { Decoys  } & Usage \\
    \midrule
    \text { CASP 10 } & 103 &  26254 & Train \\
    \text { CASP 11 } & 85 & 12563 & Train \\
    \text { CASP 12 } & 40 & 6924 & Test \\
    \text { CASP 13 } & 82 & 12336 & Train \\
    \bottomrule
    \end{tabular}
    \label{tab:dataset_casp}
}
\end{table}
\paragraph{Dataset and settings}
We use the data from past years’ editions of CASP, including CASP10-13. We randomly split the CASP10, CASP11, CASP13 for training and validation, with ratio training: validation = 9:1. CASP 12 is set aside for testing against other methods. More details about the datasets can be found in Table \ref{tab:dataset_casp}.
For the baseline, we compare our methods with other start-of-the-art methods, including random walk-based methods : RWplus~\citep{RWplus}, sequence-based methods: AngularQA~\citep{AngularQA}, and 3D structrue-based methods: VOroMQA~\citep{VoroMQA}, 3DCNN~\citep{3DCNN}. The results of these baselines we reused from ~\citet{baldassarre2020graphqa} reports. Another baseline HGNN~\citep{hgnn} is reproduced by us with same training strategy as our methods.

Because our hypergraph based methods can jointly learn node and graph embeddings, the losses in Eq. \Eqref{Eq:QAloss} can be weighted as $\mathcal{L}_{total}=\mu\mathcal{L}_l+(1-\mu)\mathcal{L}_g$ and  co-optimized by Adam Optimizer with $ L_2$ regularization, where $\mu=0.5$ in our experiment. We use grid search to select hyper-parameters and more details can be found in Table \ref{tab:hyper-parameter-QA12}.

In addition, we use CASP10-12 for training and valuation, CASP 13 for testing to further evaluate the efficiency of our methods~(the range of hyper-parameters can see Table \ref{tab:hyper-parameter-QA13}), and the results are shown in  Table \ref{tab:Protein_QA_fold_12_13}. 

\paragraph{\textcolor{blue}{Results of Protein QA.}}
Table \ref{tab:Protein_QA_fold_12_13} shows that our methods outperform most of SOTAs, which demonstrates our methods can learn protein more efficiently. The outstanding performance of H-SSGC indicates that it is able to jointly learn better the node and graph level representation. But H-SSGC trained to predict only global scores obtains $R=$ 0.712, which suggests that the local information can help the assessment of the global quality.

\subsection{Over-Smoothing Analysis.}\label{sec:Over-Smoothing_analysis}
Table \ref{tab:depth_accuracy} reveal that HyperGCN, HGNN and our H-GCN all suffer from severe over-smoothing issue, limiting the power of neural network to capture high-order relationships.

The results show that H-SSGC and H-GCN inherits the properties of SSGC ~\cite{zhu2021simple} and GCN~\cite{kipf2016semi} when the model layers are deep. H-GCN suffer from over-smoothing issue while H-SSGC significantly alleviates the performance descending with the increase of layers. Given the same layers, it can be observed that our H-SSGC almost outperforms the other methods for all cases, especially at a deep layer, which demonstrates the benefits of deep model and the long-range information around hypergraph. Furthermore, It should be noted that the optimal layer numbers $K$ of our H-SSGC are generally larger than other models, due to the only one linear layer avoids the over-fitting problem. 
\begin{table}[htbp]
\def\p{$\pm$} 
\centering
\setlength\tabcolsep{6pt}
\caption{Summary of classification accuracy (\%) results with various depths.  In our H-SSGC, the number of layers is equivalent to $K$ in \eqref{HSSGC}. We report mean test accuracy over 10 train-test splits. }
    \vspace{0mm}
\scalebox{0.85}{
\begin{tabular} {p{1.8cm}p{2.2cm}|p{0.8cm}p{0.8cm}p{0.8cm}p{0.8cm}p{0.8cm}p{0.8cm}}
\toprule
\multirow{2}{*}{Dataset} & \multirow{2}{*}{Method} & \multicolumn{6}{c}{Layers}  \\
& & 2     & 4     & 8     & 16    & 32    & 64       \\

\midrule
\multirow{5}{*}{\begin{tabular}[c]{@{}l@{}}Cora~\\(co-authorship)\end{tabular}} 
& HyperGCN\Done  & {60.66} & 57.50 & 31.09  &  31.10  & 30.09   &  31.09 \\
& HGNN\Done &{69.23} & 67.23 & 60.17 & 29.28 & 27.15 &  26.62    \\
& H-GCN (ours)\Done &\textbf{74.79} & 72.86 & 63.99 & 31.03 & 30.46 &  31.09   \\
& H-SSGC (ours)\Done & 74.60 & \textbf{75.78} & \textbf{75.70} & \textbf{75.04} & \textbf{75.26} & \textbf{74.79} \\
\midrule
\multirow{5}{*}{\begin{tabular}[c]{@{}l@{}}DBLP~\\(co-authorship)\end{tabular}}   
& HyperGCN\Done & {84.82} & 54.65 & 22.37 & 23.96 &  23.04  & 24.13      \\
& HGNN\Done & {88.55}& 88.28 & 85.38 & 27.64 &27.62 & 27.56 \\
& H-GCN (ours)\Done &\textbf{89.04} & \textbf{88.90} & 85.15 & 27.61 & 27.61 &  27.62   \\
& H-SSGC (ours)\Done & 86.63 & 88.26 & \textbf{89.00}& \textbf{89.17} & \textbf{89.05} &  \textbf{88.60} \\
\midrule
\multirow{5}{*}{\begin{tabular}[c]{@{}l@{}}Cora~\\(co-citation)\end{tabular}}     
& HyperGCN\Done & 62.35 & 58.29 & 31.09  & 31.17 & 31.09  & 29.68  \\
& HGNN\Done & {55.60}& 55.72& 42.10 & 26.16 & 24.40 & 24.43    \\
& H-GCN (ours)\Done &\textbf{69.03} & \textbf{69.45} & 57.37 & 28.21 & 26.27 &  26.95   \\
& H-SSGC (ours)\Done & 62.21 & 64.57 & \textbf{67.59} & \textbf{68.96} & \textbf{69.37} & \textbf{68.15} \\
\midrule
\multirow{5}{*}{\begin{tabular}[c]{@{}l@{}}Pubmed~\\(co-citation)\end{tabular}}   
& HyperGCN\Done & 68.12 & 63.59 & 39.99 & 39.97  & 40.01  & 40.02   \\
& HGNN\Done & {46.41} & 47.16 & 40.93 & 40.24 & 40.30 & 40.29    \\
& H-GCN (ours)\Done &\textbf{75.37} & {74.76} & 60.65 & 40.38 & 40.31 &  40.42   \\
& H-SSGC (ours)\Done & 74.39 & \textbf{74.91} & \textbf{74.41} & \textbf{73.90} & \textbf{72.79} & \textbf{71.49}  \\
\midrule
\multirow{5}{*}{\begin{tabular}[c]{@{}l@{}}Citeseer~\\(co-citation)\end{tabular}} 
& HyperGCN\Done & 56.94  & 36.75 & 20.72 & 20.41  & 20.16  & 18.95     \\
& HGNN\Done & {39.93 }& 38.98 & 36.67 & 19.91 &  19.86 & 19.79    \\
& H-GCN (ours)\Done &\textbf{62.67} & 61.50 & 49.94 & 21.95 & 21.84 &  21.93   \\
& H-SSGC (ours)\Done & 61.63 & \textbf{62.75} & \textbf{63.86} & \textbf{64.62} & \textbf{65.14} & \textbf{65.10}    \\
\bottomrule

\end{tabular}
}

    \vspace{0mm}
\label{tab:depth_accuracy}
\end{table}

\subsection{Ablation Analysis}\label{sec:Ablation_analysis}

\begin{table}[thbp]
\def\p{$\pm$} 
\centering
\setlength\tabcolsep{10pt} 
\caption{Test accuracy~(\%) and standard deviation for 10 random seeds. {w/o $\Q$} represents replacing EDVWs $\Q$ with EIVWs $\mbf{H}$. (p-value is gained by the t-test) }
\scalebox{1}{
\begin{tabular}{l|l|ll}
    \toprule 
    \multirow{1}{*} { Methods }& \multirow{1}{*} {  } & NTU & ModelNet40 \\ 
    \hline
    \multirow{3}{*}{\begin{tabular}[c]{@{}l@{}}H-GCN\end{tabular}} 
    & w/o $\Q$  &  84.93\p0.31  &   97.18\p0.20  \\
    & w/ $\Q$ & {85.15\p0.34}$\uparrow_{0.78}$ & {97.28\p 0.15}$\uparrow_{0.1}$ \\ 
    \cline{2-4}
    & p-value &  0.0739 & 0.1110 \\
    \hline
    \multirow{3}{*}{\begin{tabular}[c]{@{}l@{}}H-SSGC\end{tabular}} 
    & w/o $\Q$ & 83.03$\pm$0.30 & 96.82$\pm$0.10 \\
    & w/ $\Q$ & 84.13$\pm$0.29$\uparrow_{1.1}$ & 96.94$\pm$0.06$\uparrow_{0.12}$  \\ 
    \cline{2-4}
    & p-value & 5.63e-8 & 0.0015\\
    \hline
    \multirow{3}{*}{\begin{tabular}[c]{@{}l@{}}H-GCNII\end{tabular}} 
    & w/o $\Q$       &   85.04$\pm$0.29 & 97.62$\pm$0.07  \\
    & w/ $\Q$       & 85.17$\pm$0.36$\uparrow_{0.13}$  & 97.75$\pm$0.07$\uparrow_{0.13}$   \\
    \cline{2-4}
    & p-value &  0.1928 & 0.0003 \\
    \bottomrule
\end{tabular}
}
    \label{tab:ab_Q}    
\end{table}
\textbf{\textcolor{blue}{Effect of EDVWs.}}
We conduct an ablation study to verify that our framework is expert in  EDVW-hypergraph learning. The results in Tabel \ref{tab:ab_Q} indicate that our methods can make use of the EDVWs to achieve better performance.
Meanwhile, 'w/ $\Q$' outperforms 'w/o $\Q$' significantly, indicating that our H-GNNs models successfully mine the information in the vertex weights.
\begin{table*}[thbp]
\vspace{-2mm}
\def\p{$\pm$} 
\caption{classification accuracy $\pm$ standard deviation) with different $\rho(x)$ used in Eq. \eqref{eq:h-gnns}. ($\phi(x)$ is Gaussian PDF.)
}
\setlength\tabcolsep{6pt} 
\centering
\vspace{-3mm}
\scalebox{0.765}{
    \begin{tabular}{c|lccccccccc}
    \toprule
    Dataset & \text{Methods} & \text{Random}	& $x^{-1}$	& $x^{0}$ & $x^{1}$ &  $\operatorname{sigmoid}(x)$	& $\phi(x)$ &	$\log(x)$ & $\exp(x)$ &	$\exp(-x)$ \\
    \midrule
    \text{coauthorship/cora} &	\text{H-APPNP} & 75.21$\pm$0.6 & \textbf{76.38$\pm$0.8} & 75.51$\pm$1.0 & 75.06$\pm$0.9 & 75.49$\pm$1.0 & 75.78$\pm$0.6 & 75.18$\pm$0.9 &  74.16$\pm$0.8 & 64.6$\pm$1.4 \\
    \text{cocitation/pubmed} &	\text{H-APPNP} & 75.43$\pm$1.0 & 75.31$\pm$1.1  &  {75.46$\pm$1.0} & 75.29$\pm$1.2 & 75.45$\pm$1.0 & \textbf{75.47$\pm$1.1} & 75.40$\pm$1.1 & 74.44$\pm$1.4 & 73.54$\pm$1.3\\
    
    \text{coauthorship/cora} &	\text{H-GCNII} &	75.68$\pm$0.8 & \textbf{76.21$\pm$1.0} & 75.57$\pm$0.8 &75.39$\pm$0.8  & 75.64$\pm$1.0	&76.15$\pm$0.9	 &  75.29$\pm$0.8& 74.64$\pm$0.9 & 65.47$\pm$1.0 \\
    \text{cocitation/pubmed} & \text{H-GCNII}	& 75.50$\pm$1.2 & \textbf{75.79$\pm$1.1} &75.74$\pm$1.3 & 74.85$\pm$1.4 & 75.61$\pm$1.4 &	74.64$\pm$0.9 &	75.39$\pm$1.3 & 73.62$\pm$1.9	 &74.68$\pm$1.2  \\
    \bottomrule
    \end{tabular}
\label{tab:analysisOfRho}
}
\vspace{-5mm}
\end{table*}


\textbf{\textcolor{blue}{Effect of Different $\rho$.}}
We investigate the performance of the proposed method with different $\rho$ on citation
datasets. As illustrated in Table~\ref{tab:analysisOfRho}, H-GNNs have different performances under various $\rho$, indicating the $\rho$ can influence the performance of neural networks by the random walk. Meanwhile, the $x^{-1}$ achieves a satisfactory performance. This can be explained as follows.
In co-authorship graphs, papers with a larger number of co-authors are often the result of collaborations between researchers from various fields, and therefore will not be as predictive as papers with fewer authors. The co-citation graph enjoys a similar explanation.
\vspace{-1mm}

\paragraph{\textcolor{blue}{Effect of Re-normalization Trick.}} In order to verify the effectiveness of our proposed edge-dependent vertex weight $\Q$ and the necessity of re-normalization trick, we conduct an ablation analysis and report the results in Table \ref{tab:ab_study}. Specially, \textit{w/o $\Q$} is a variant of our methods that replaces the edge-dependent vertex weight $\Q$ with the edge-independent vertex weight matrix $\H$. \textit{w/o renormalization} represents  the variants without re-normalization trick. Form the reported results we can learn that both $\Q$ and renormalization are efficient for hypergpraph learning, respectively. Moreover, on Cora and Pubmed, the performance gap between \textit{w/o renormalization} and \textit{with both} H-GCN reveals the 
advantage of renormalization on disconnected hypergraph dataset. This phenomenon is mainly caused by the row in the adjacent matrix corresponding to an isolated point is 0~(Figure \ref{fig:hypergraph2graph}), resulting in a direct loss of its vertex information~(Figure \ref{fig:hypergraph2graph}). And the renormalization trick adding the self-loop to matrix $\K$ can maintain the features of isolated vertices during aggregation. Compared with H-GCN, the performance impact of renormalization on H-SSGC seems to be smaller. This is because we add initial features to the information gathered by the neighbors, which reduces the information loss of isolated vertices.

\begin{table*}[thbp]
\def\p{$\pm$} 
\centering
\setlength\tabcolsep{4pt} 
\caption{Test accuracy~(\%) of our methods for ablation analysis. We report mean \p standard deviation. Cora and Pubmed are the datasets that do not contain $\Q$, so we just report the w/o renormalization results.  NTU and ModelNet40 constructed by \citet{hgnn} are both connected hypergraph networks in this work. }
    \vspace{0mm}
\scalebox{0.88}{
\begin{tabular}{c|l|cc|cc}
    \toprule 
    \multirow{1}{*} { methods }& \multirow{1}{*} { - } & Cora (co-authorship)  & Pubmed (co-citation) & NTU & ModelNet40 \\
    \hline
    \multirow{4}{*}{\begin{tabular}[c]{@{}l@{}}H-GCN\end{tabular}} 
    & with both & \textbf{74.79\p0.91}  & \textbf{75.37\p1.2}  &  \textbf{85.15\p0.34}  &   \textbf{97.28\p 0.15}   \\
    & w/o $\Q$ & -    &  -   & 84.93\p0.31 &  97.18\p0.20   \\
    & w/o renormalization & 69.23\p 1.6  & 46.41\p 0.70  &  84.85\p 0.30  &   97.20\p0.15   \\
    & w/o both & -  & -  &  84.21\p 0.25  &   97.15\p0.14   \\
    
    \hline
    \multirow{4}{*}{\begin{tabular}[c]{@{}l@{}}H-SSGC\end{tabular}} 
    & with both      &    \textbf{75.91\p 0.75}     & \textbf{74.60\p1.4}   &   \textbf{83.35\p 0.30} & \textbf{97.74\p 0.05}\\
    & w/o $\Q$       &      -   &  -  & 82.84\p 0.40  & 97.65\p0.08  \\
    & w/o renormalization &  75.76\p0.69 &  73.83\p 2.1 & 82.92\p 0.47   &   \textbf{97.74\p 0.05}\\
    & w/o both &  - &  - & 82.55\p 0.46   &   {97.69\p 0.07} \\
    \bottomrule
\end{tabular}
}
    \vspace{0mm}
    \label{tab:ab_study}    
\end{table*}

\subsection{Running Time and Computational Complexity}\label{appsec:compute_complexity}

The H-GNNs has same theoretical computational complexity as the undigraph NNs. For example,
the computational cost of H-GCN is$\mathcal{O}(|E|d)$, where $|E|$ is the total edge count in equivalent undigraph. Each sparse matrix multiplication $\tilde{\T}\mathbf{X}$ costs $|E|d$. And the computational of HGNN is the same as H-GCN.

Then, we compare the running time of our H-GNNs with existing models in Table \ref{tab:runningTime} and the results illustrate that our methods are of the same order of magnitude as SOTA's approach UniGNN and outperform the HyperGCN and HGAT.

\begin{table}[htbp]

\centering
\def\p{$\pm$} 
\caption{The average training time per epoch with different methods on citation network classification task is shown below and timings are measured in seconds. The float in parentheses is the standard deviation.}
\scalebox{0.8}{

    \begin{tabular}{c|c|c|c|c|c}
    \toprule
     \text{Methods}  & \text{cora coauthorship} & \text{dblp coauthorship} & \text{cora cocitation} & \text{pubmed cocitation} & \text{citeseer cocitation} \\
    \midrule
    \text{HyperGCN} & 0.150\p0.058 & 1.181\p0.071 & 0.151\p0.029& 1.203\p0.104	& 0.130\p0.029 \\
    \text{HGNN} & 0.005\p0.002 & 0.081\p0.006 & 0.005\p0.040 & 0.008\p0.002 & 	0.005\p0.002\\
    \text{UniGNN } & 0.014\p0.044 & 0.042\p0.040 & 0.014\p0.042 & 0.023\p0.043 & 0.0168\p0.043\\
    \text{HNHN} & 0.001\p0.0026 & 0.007\p0.014 & 0.0010\p0.004 & 0.009\p0.006 &  0.001\p0.003 \\
    HGAT & 0.381\p0.080 & OOM & 0.279\p0.083 & 1.329\p0.016 &0.286\p0.087 \\
    \midrule
    \text{H-ChebNet} & 0.027\p0.005& 0.063\p0.001 & 0.073\p0.008 &0.050\p0.019 & 0.067\p0.0198\\
    H-SSGC & 0.055\p0.001 & 0.291\p0.001 & 0.205\p0.003 & 0.135\p0.056 & 0.193\p0.057 \\
    \text{H-GCN} & 0.005\p0.036& 0.020\p0.079& 0.005\p0.039 & 0.011\p0.075 &  0.081\p0.091\\
    H-APPNP & 0.147$\pm$0.013 & 1.017$\pm$0.080 &0.033$\pm$0.006 & 0.319$\pm$0.037 & 0.123$\pm$0.011 \\
    H-GCNII & 0.141$\pm$0.01 & 0.252$\pm$0.017 & 0.135$\pm$0.013 & 0.121$\pm$0.014 & 0.079$\pm$0.004 \\
    \bottomrule
    \end{tabular}
}

\label{tab:runningTime}
\end{table}

\subsection{Comparison between edge-dependent vertex weights and hypergraph attention }
\begin{figure*}[htbp]
\centering
    \subfigure{
    \begin{minipage}[s]{0.3\linewidth}
    \centering
    \includegraphics[width=1\linewidth]{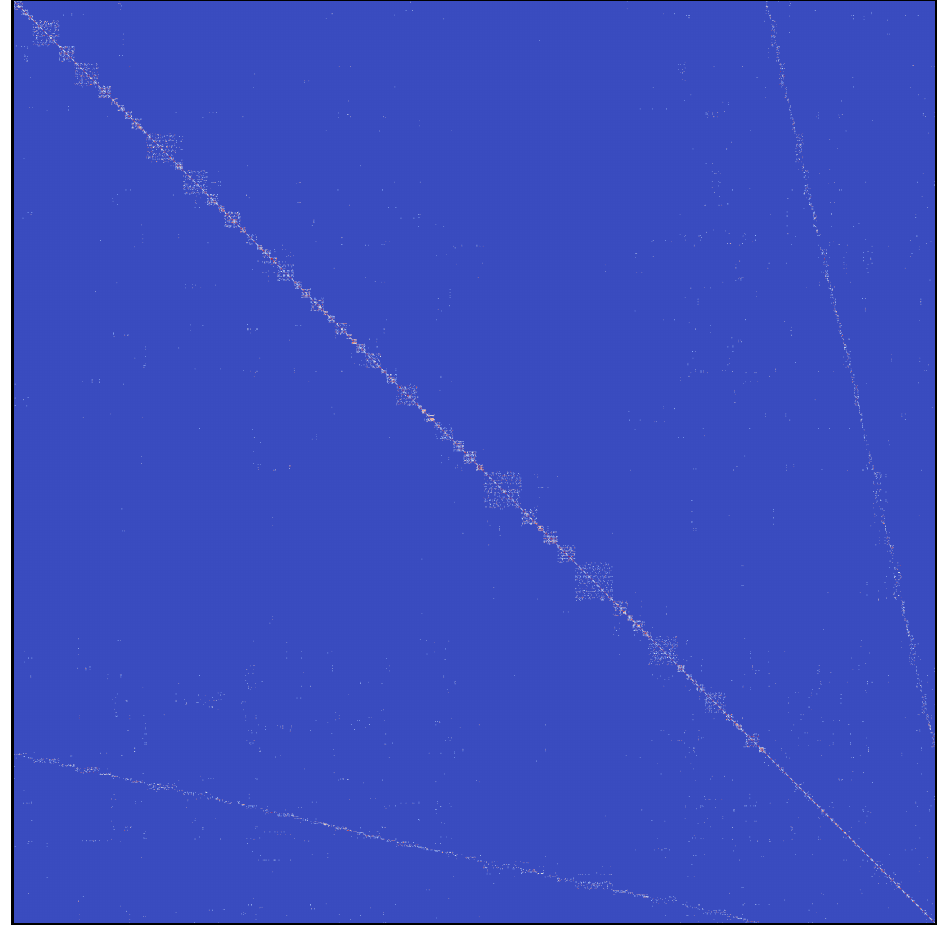} 
    \end{minipage}%
    }%
    \quad 
    \subfigure{
    \begin{minipage}[s]{0.3\linewidth}
    \centering
    \includegraphics[width=1\linewidth]{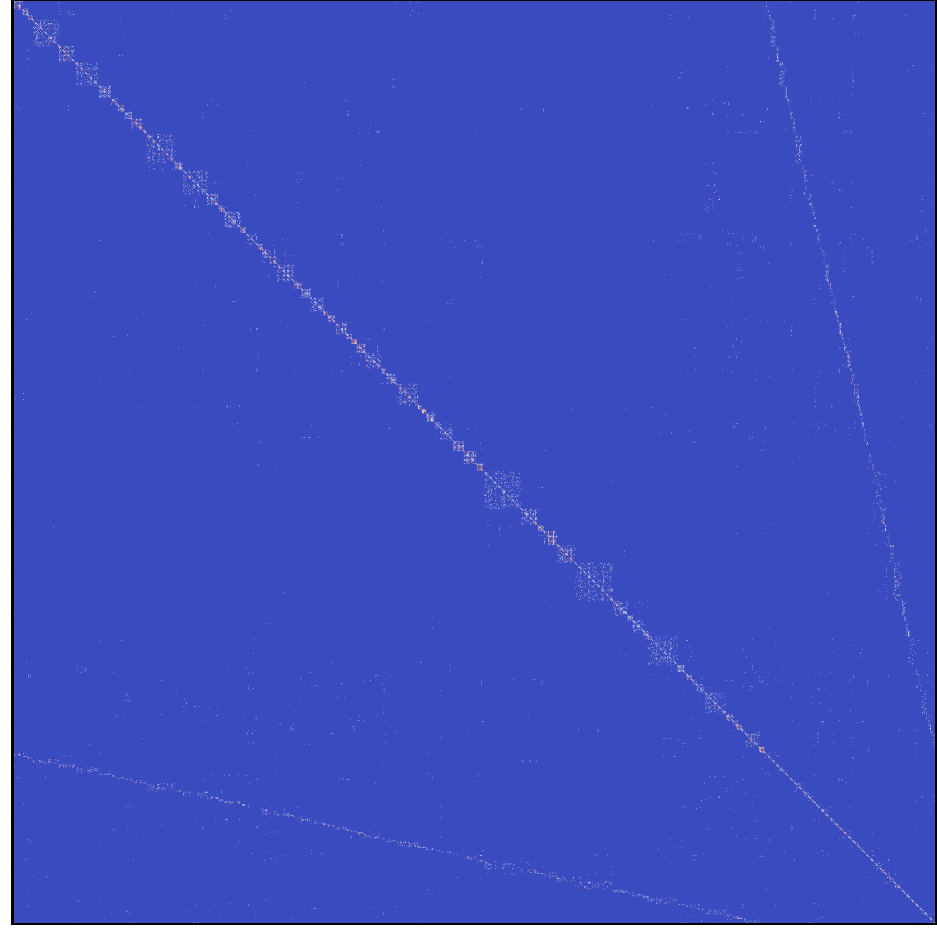}
    \end{minipage}%
    }%
    \quad
    \subfigure{
    \begin{minipage}[s]{0.3\linewidth}
    \centering
    \includegraphics[width=1\linewidth]{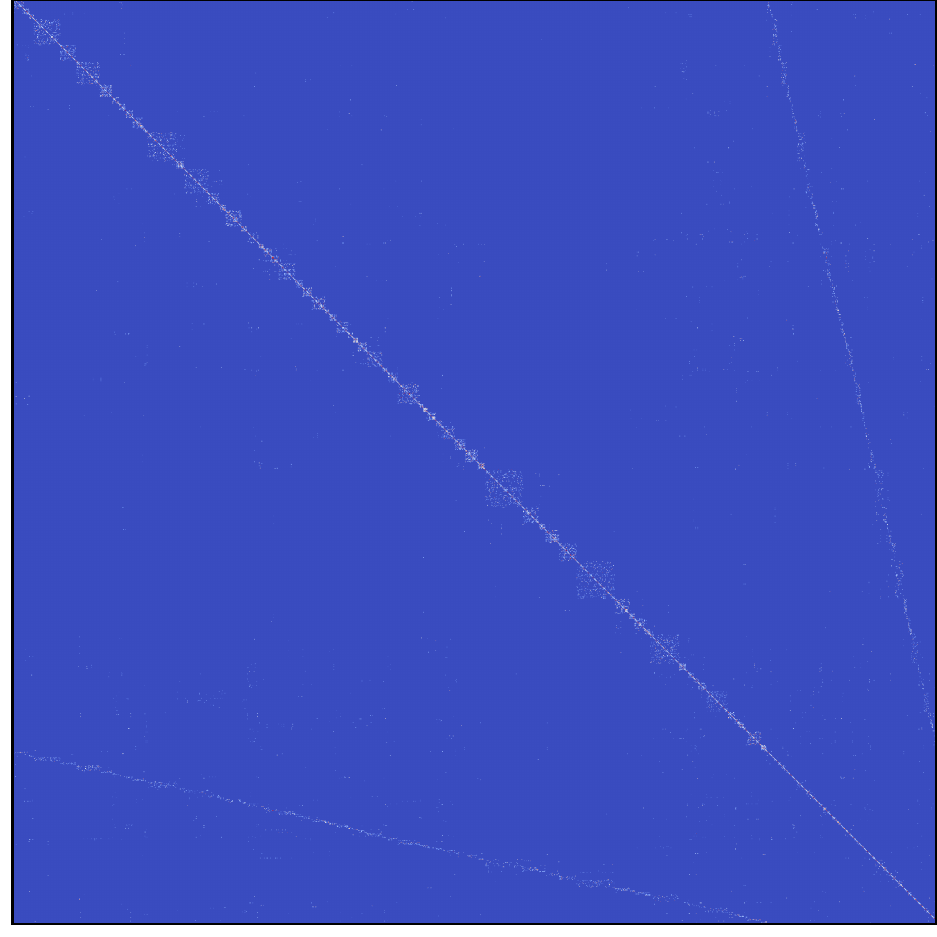}
    \end{minipage}
    }%

\centering

\caption{The edge-dependent vertex weights of NTU2012 dataset~(MVCNN feature + MVCNN structure).  The figure on the left represents the distance-based vertex weights matrix $\Q~$ used in H-GCN. 
The middle~($\Q_1$~) and right~($\Q_2$) denote the node-level and edge-level attention coefficient matrix of HGAT~\citep{HyperGAT}, respectively.}
\label{fig:attention_map}
\end{figure*}

The node-level and edge-level attention in HGAT~\cite{HyperGAT} can be considered as the learnable EDVWs.
we visualize EDVWs in Figure \ref{fig:attention_map}.

\subsubsection{Visual Object.}

\end{document}